\documentclass{article}

\PassOptionsToPackage{numbers,compress}{natbib}

\usepackage[final]{neurips_2021}

\usepackage[utf8]{inputenc} 
\usepackage[T1]{fontenc}    
\usepackage{hyperref}       
\usepackage{url}            
\usepackage{booktabs}       
\usepackage{amsfonts}       
\usepackage{nicefrac}       
\usepackage{microtype}      
\usepackage{xcolor}         

\usepackage{booktabs}       
\usepackage{amsfonts}       
\usepackage{nicefrac}       
\usepackage{microtype}      
\usepackage{dsfont}
\usepackage{amsmath}
\usepackage{graphicx}
\usepackage{subfig}
\usepackage{bm}
\usepackage{amsthm}
\usepackage{comment}
\usepackage{multirow}
\usepackage{floatrow}
\newtheorem{proposition}{Proposition}

\newfloatcommand{capbtabbox}{table}[][\FBwidth]
\usepackage{xr}
\makeatletter
\newcommand*{\addFileDependency}[1]{
  \typeout{(#1)}
  \@addtofilelist{#1}
  \IfFileExists{#1}{}{\typeout{No file #1.}}
}
\makeatother

\newcommand*{\myexternaldocument}[1]{%
    \externaldocument{#1}%
    \addFileDependency{#1.tex}%
    \addFileDependency{#1.aux}%
}

\title{DSelect-k: Differentiable Selection in the Mixture of Experts with Applications to Multi-Task Learning}

\author{%
Hussein Hazimeh\textsuperscript{1}, Zhe Zhao\textsuperscript{1}, Aakanksha Chowdhery\textsuperscript{1}, Maheswaran Sathiamoorthy\textsuperscript{1} \\[0.5em] \textbf{Yihua Chen\textsuperscript{1}}, \textbf{Rahul Mazumder\textsuperscript{2}}, \textbf{Lichan Hong\textsuperscript{1}}, \textbf{Ed H. Chi\textsuperscript{1}}\\ \\
  \textsuperscript{1}Google, \texttt{\{hazimeh,zhezhao,chowdhery,nlogn,yhchen,lichan,edchi\}@google.com}\\
  \textsuperscript{2}Massachusetts Institute of Technology, \texttt{rahulmaz@mit.edu}\\
}

\myexternaldocument{supp_camera_ready}

\begin{document}

\maketitle

\begin{abstract}
The Mixture-of-Experts (MoE) architecture is showing promising results in improving parameter sharing in  multi-task learning (MTL) and in scaling high-capacity neural networks. State-of-the-art MoE models use a trainable ``sparse gate'' to select a  subset of the experts for each input example. While conceptually appealing, existing sparse gates, such as Top-k, are not smooth. The lack of smoothness can lead to convergence and statistical performance issues when training with gradient-based methods. In this paper, we develop DSelect-k: a continuously differentiable and sparse gate for MoE, based on a novel binary encoding formulation. The gate can be trained using first-order methods, such as stochastic gradient descent, and offers explicit control over the number of experts to select. We demonstrate the effectiveness of DSelect-k on both synthetic and real MTL datasets with up to $128$ tasks. Our experiments indicate that DSelect-k can achieve statistically significant improvements in prediction and expert selection over popular MoE gates. Notably, on a real-world, large-scale recommender system, DSelect-k achieves over $22\%$  improvement in predictive performance compared to Top-k. We provide an open-source implementation of DSelect-k\footnote{\url{https://github.com/google-research/google-research/tree/master/dselect_k_moe}}.
\end{abstract}

\section{Introduction}
\begin{figure}[htbp]
    \centering
    \includegraphics[scale=0.21]{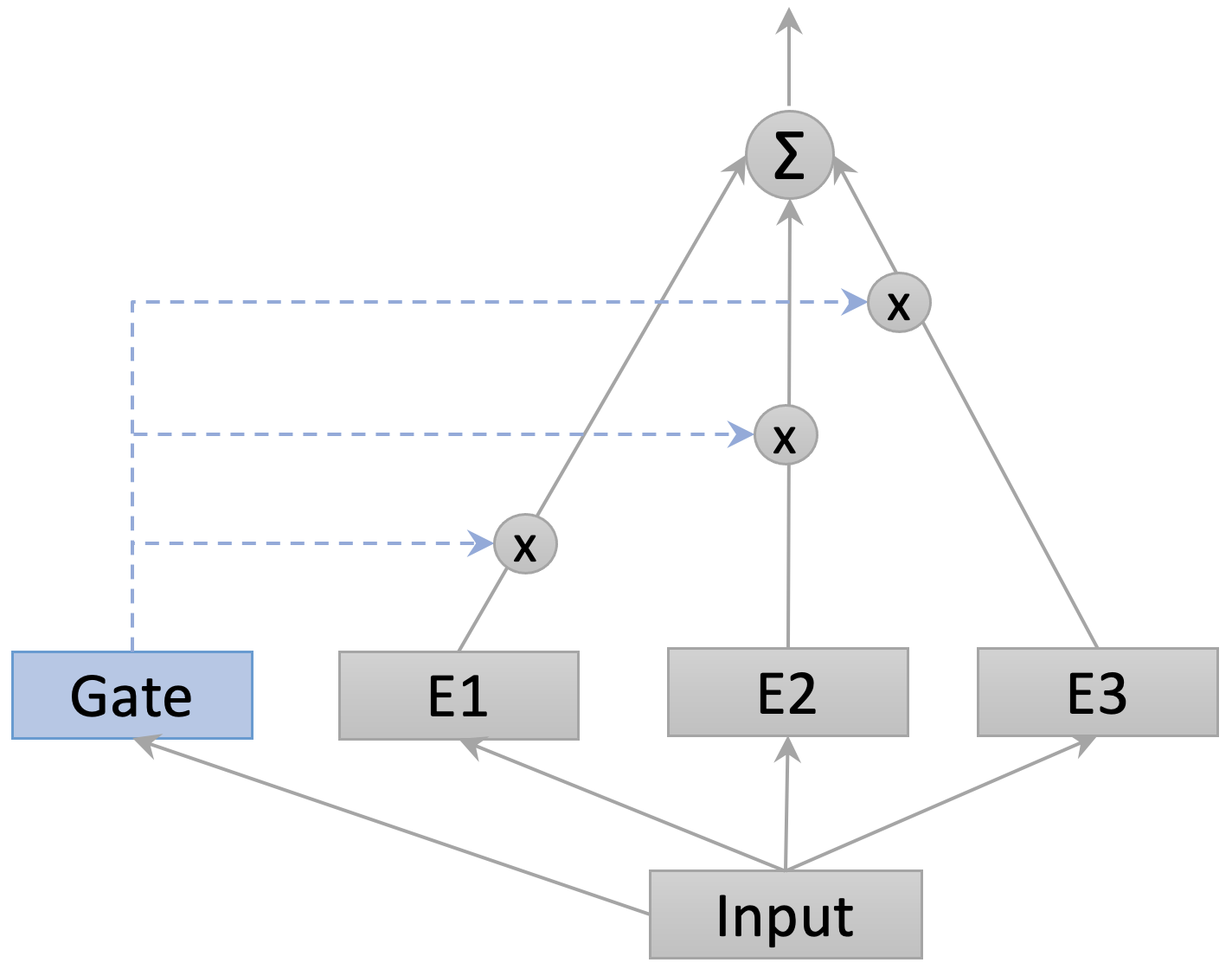} \qquad
    \includegraphics[scale=0.21]{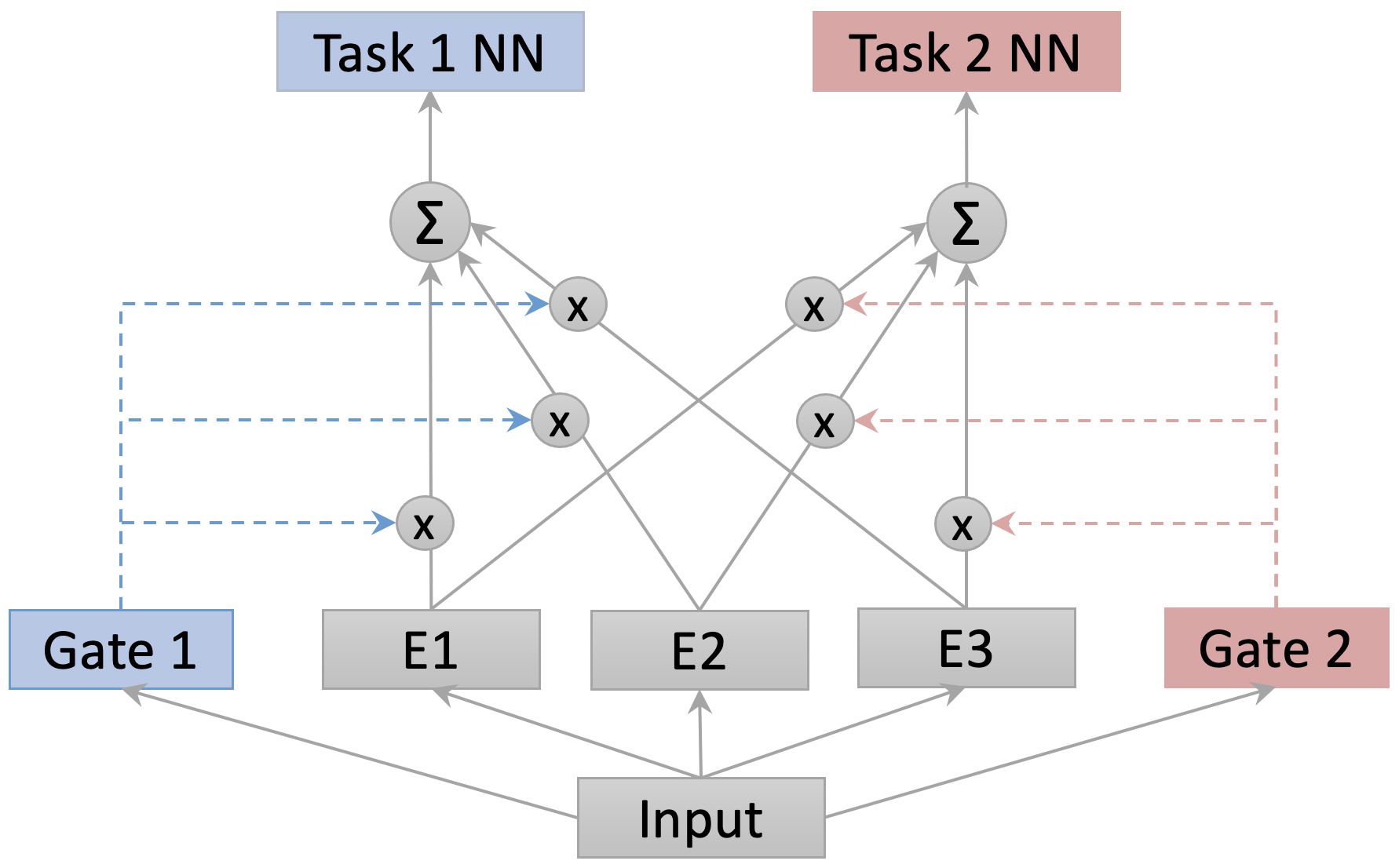}
    \caption{\textbf{(Left)}: An example of a MoE that can be used as a standalone learner or layer in a neural network. Here ``Ei'' denotes the $i$-th expert. \textbf{(Right):} A multi-gate MoE for learning two tasks simultaneously. ``Task i NN'' is a neural network that generates the output of Task i. }
    \label{fig:MoE}
\end{figure}

The Mixture of Experts (MoE) \cite{jacobs1991adaptive} is the basis of many state-of-the-art deep learning models. For example, MoE-based layers are being used to perform efficient computation in high-capacity neural networks and to improve parameter sharing in multi-task learning (MTL) \cite{shazeer2017outrageously,ma2018modeling,lepikhin2020gshard}. 
\textcolor{black}{In its simplest form, a MoE consists of a set of experts (neural networks) and a trainable gate. The gate assigns weights to the experts on a per-example basis, and the MoE outputs a weighted combination of the experts. This per-example weighting mechanism allows experts to specialize in different partitions of the input space, which has the potential to improve  predictive performance and interpretability. In Figure \ref{fig:MoE} (left), we show an example of a simple MoE architecture that can be used as a standalone learner or as a layer in a neural network.}


The literature on the MoE has traditionally focused on softmax-based gates, in which all experts are assigned nonzero weights \citep{jordan1994hierarchical}. \textcolor{black}{To enhance the computational efficiency and interpretability of MoE models, recent works use \textsl{sparse gates} that assign nonzero weights to only a small subset of the experts \citep{DBLP:journals/corr/BengioBPP15,shazeer2017outrageously,rosenbaum2018routing,lepikhin2020gshard}.} Existing sparse gates are not differentiable, and reinforcement learning algorithms are commonly used for training \cite{DBLP:journals/corr/BengioBPP15,rosenbaum2018routing}. In an exciting work, \cite{shazeer2017outrageously} introduced a new sparse gate (Top-k gate) and proposed training it using stochastic gradient descent (SGD). The ability to train the gate using SGD is appealing because it enables end-to-end training. However, the Top-k gate is not continuous, which can lead to convergence issues in SGD that affect statistical performance (as we demonstrate in our experiments).

In this paper, we introduce DSelect-k: a continuously differentiable and sparse gate for MoE. Given a user-specified parameter $k$, the gate selects at most $k$ out of the $n$ experts. This explicit control over sparsity leads to a cardinality-constrained optimization problem, which is computationally challenging. To circumvent this challenge, we propose a novel, unconstrained reformulation that is equivalent to the original problem. The reformulated problem uses a binary encoding scheme to implicitly enforce the cardinality constraint. We demonstrate that by carefully  smoothing the binary encoding variables, the reformulated problem can be effectively optimized using first-order methods such as SGD. \textcolor{black}{DSelect-k has a unique advantage over existing methods in terms of compactness and computational efficiency. The number of parameters  used by DSelect-k is logarithmic in the number of experts, as opposed to linear in existing gates such as Top-k. Moreover, DSelect-k's output can be computed efficiently via a simple, closed-form expression. In contrast, state-of-the-art differentiable methods for stochastic k-subset selection and Top-k relaxations, such as  \cite{paulus2020gradient,xie2020differentiable}\footnote{These methods were not designed specifically for the MoE.}, require solving an optimization subproblem (for each input example) to compute the gate's output.}

DSelect-k supports two gating mechanisms: \textsl{per-example} and \textsl{static}. Per-example gating is the classical gating technique used in MoE models, in which the weights assigned to the experts are a function of the input example \citep{jacobs1991adaptive, shazeer2017outrageously}. In static gating, a subset of experts is selected and the corresponding weights do not depend on the input \citep{rosenbaum2018routing}. 
Based on our experiments, each gating mechanism can outperform the other in certain settings. Thus, we study both mechanisms and advocate for experimenting with each. 

MTL is an important area where MoE models in general, and our gate in particular, can be useful. The goal of MTL is to learn multiple tasks simultaneously by using a shared model. Compared to the usual single task learning, MTL can achieve better generalization performance through exploiting task relationships  \citep{caruana1997multitask}. One key problem in MTL is how to share model parameters between tasks \citep{ruder2017overview}. For instance, sharing parameters between unrelated tasks can potentially degrade performance. The multi-gate MoE \citep{ma2018modeling} is a flexible architecture that allows for learning what to share between tasks. Figure \ref{fig:MoE} (right) shows an example of a multi-gate MoE, in the simple case of two tasks. Here, each task has its own gate that adaptively controls the extent of parameter sharing. In our experiments, we study the effectiveness of DSelect-k in the context of the multi-gate MoE.


\textbf{Contributions: } On a high-level, our main contribution is DSelect-k: a new continuously differentiable and sparse gate for MoE, which can be directly trained using first-order methods. Our technical contributions can be summarized as follows. \textbf{(i)} The gate selects (at most) $k$ out of the $n$ experts, where $k$ is a user-specified parameter. This leads to a challenging, cardinality-constrained optimization problem. To deal with this challenge, \textcolor{black}{we develop a novel, unconstrained reformulation, and we prove that it is equivalent to the original problem.} The reformulation uses a binary encoding scheme that implicitly imposes the cardinality constraint using learnable binary codes. \textbf{(ii)} To make the unconstrained reformulation smooth, we relax and smooth the binary variables. We demonstrate that, with careful initialization and regularization, the resulting problem can be optimized with first-order methods such as SGD. \textbf{(iii)} We carry out a series of experiments on synthetic and real MTL datasets, which show that our gate is \textcolor{black}{competitive with}  state-of-the-art gates in terms of parameter sharing and predictive performance. \textbf{(iv)} We provide an open-source implementation of DSelect-k.

\subsection{Related Work}
\textbf{MoE and Conditional Computation: } Since MoE was introduced by \cite{jacobs1991adaptive}, an exciting body of work has extended and studied this model, e.g., see \cite{jordan1994hierarchical,jacobs1997bias,jiang1999identifiability}.
Recently, MoE-based models are showing success in deep learning. For example, \cite{shazeer2017outrageously} introduced the sparse Top-k gate for MoE and showed significant computational improvements on machine translation tasks; we discuss exact connections to this gate in Section \ref{sec:MoE}. The Top-k gate has also been utilized in several state-of-the-art deep learning models that considered MTL tasks, e.g., \cite{lepikhin2020gshard,ramachandran2018diversity,fedus2021switch}.
Our work is also related to the conditional computation models that activate parts of the neural network based on the input  \citep{bengio2013estimating,DBLP:journals/corr/BengioBPP15,shazeer2017outrageously,ioannou2016decision,wang2018skipnet}. Unlike DSelect-k, these works are based on non-differentiable models, or heuristics where the training and inference models are  different.

\textcolor{black}{\textbf{Stochastic k-Subset Selection and Top-k Relaxation}}: A related line of work focuses on stochastic k-subset selection in neural networks, e.g., see \cite{paulus2020gradient,chen2018learning,xie2019reparameterizable} and the references therein. Specifically, these works propose differentiable methods for sampling $k$-subsets from a categorical distribution, based on extensions or generalizations of the Gumbel-softmax trick \citep{maddison2016concrete,jang2016categorical}. However, in the MoE we consider \textsl{deterministic} subset selection---determinism is a common assumption in MoE models that can improve interpretability and allows for efficient implementations  \citep{jacobs1991adaptive,jordan1994hierarchical,shazeer2017outrageously}. In contrast, the stochastic approaches described above are suitable in applications where there is an underlying sampling distribution, such as in variational inference \citep{kingma2013auto}. \textcolor{black}{Another related work is the differentiable relaxation of the Top-k operator proposed by \citep{xie2020differentiable}. All the aforementioned works perform dense training (i.e., the gradients of all experts, even if not selected, will have to be computed during backpropagation), whereas DSelect-k can (to an extent) exploit sparsity to speed up training, as we will discuss in Sections \ref{sec:MoE} and \ref{sec:our_gate}. Moreover, the stochastic k-subset selection framework in \citep{paulus2020gradient} (which encompasses several previous works) and the Top-k relaxation in \cite{xie2020differentiable} require solving an optimization subproblem to compute the gate output--each example will require solving a separate subproblem in the per-example gating setting, which can be computationally prohibitive. In contrast, DSelect-k's output is computed efficiently via a closed-form expression.}

\textcolor{black}{\textbf{Sparse Transformations to the Simplex: } These are sparse variants of the softmax function that can output sparse probability vectors, e.g., see  \cite{martins2016softmax,peters2019sparse,correia2019adaptively,blondel2019learning}. While similar to our work in that they output sparse probability vectors, these transformations cannot control the sparsity level precisely as DSelect-k does (through a cardinality constraint). Thus, these transformations may assign some examples or tasks sparse combinations and others dense combinations.}

\textcolor{black}{\textbf{MTL: } In Appendix \ref{sec:appendix_related_work}, we review related literature on MTL.}
\section{Gating in the Mixture of Experts} \label{sec:MoE}
In this section, we first review the MoE architecture and popular gates, and then discuss how these gates compare to our proposal. We will assume that the inputs to the MoE belong to a space $\mathcal{X} \subset \mathbb{R}^{p}$. In its simplest form, the MoE consists of a set of $n$ experts (neural networks) $f_{i}: \mathcal{X} \to \mathbb{R}^{u}$, $i \in \{1,2,\dots, n\}$, and a gate  $g: \mathcal{X} \to \mathbb{R}^n$ that assigns weights to the experts. The gate's output is assumed to be a probability vector, i.e., $g(x) \geq 0$ and $\sum_{i=1}^n g(x)_i = 1$, for any  $x \in \mathcal{X}$. Given an example $x \in \mathcal{X}$, the corresponding output of the MoE is a weighted combination of the experts:
\begin{align} \label{eq:moe_output}
    \sum_{i=1}^n f_i(x) g(x)_i.
\end{align}
Next, we discuss two popular choices for the gate $g(.)$ that can be directly optimized using SGD.

\textbf{Softmax Gate: } A classical model for $g(x)$ is the softmax gate: $\sigma(A x + b)$, where $\sigma(.)$ is the softmax function, $A \in \mathbb{R}^{n \times p}$ is a trainable weight matrix, and $b \in \mathbb{R}^{n}$ is a bias vector \cite{jordan1994hierarchical}. This gate is dense, in the sense that all experts are assigned nonzero probabilities. Note that static gating (i.e., gating which does not depend on the input example) can be obtained by setting $A = 0$. 

\begin{figure}[tbp]
    \centering
    \includegraphics[scale=0.37]{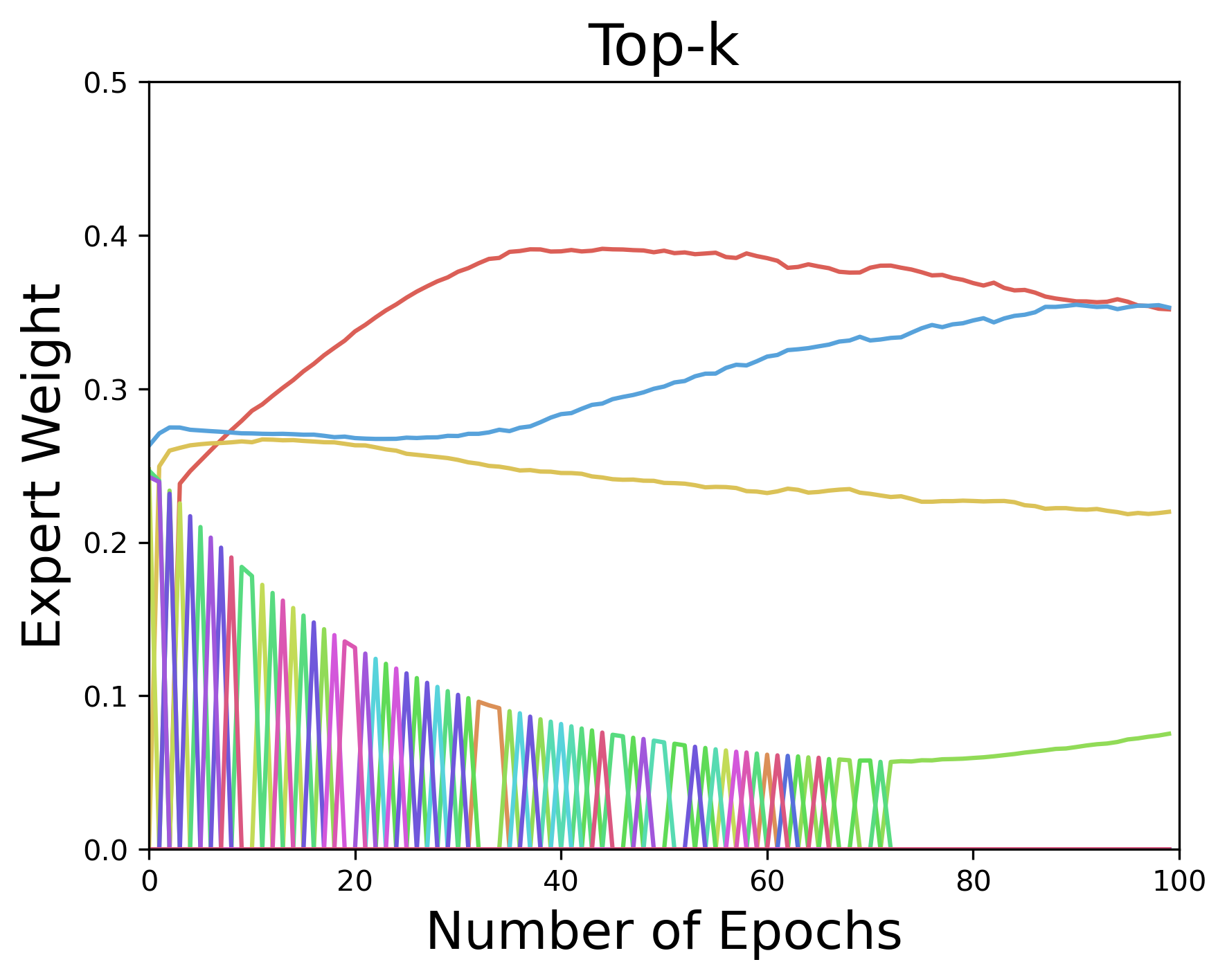} \qquad
    \includegraphics[scale=0.37]{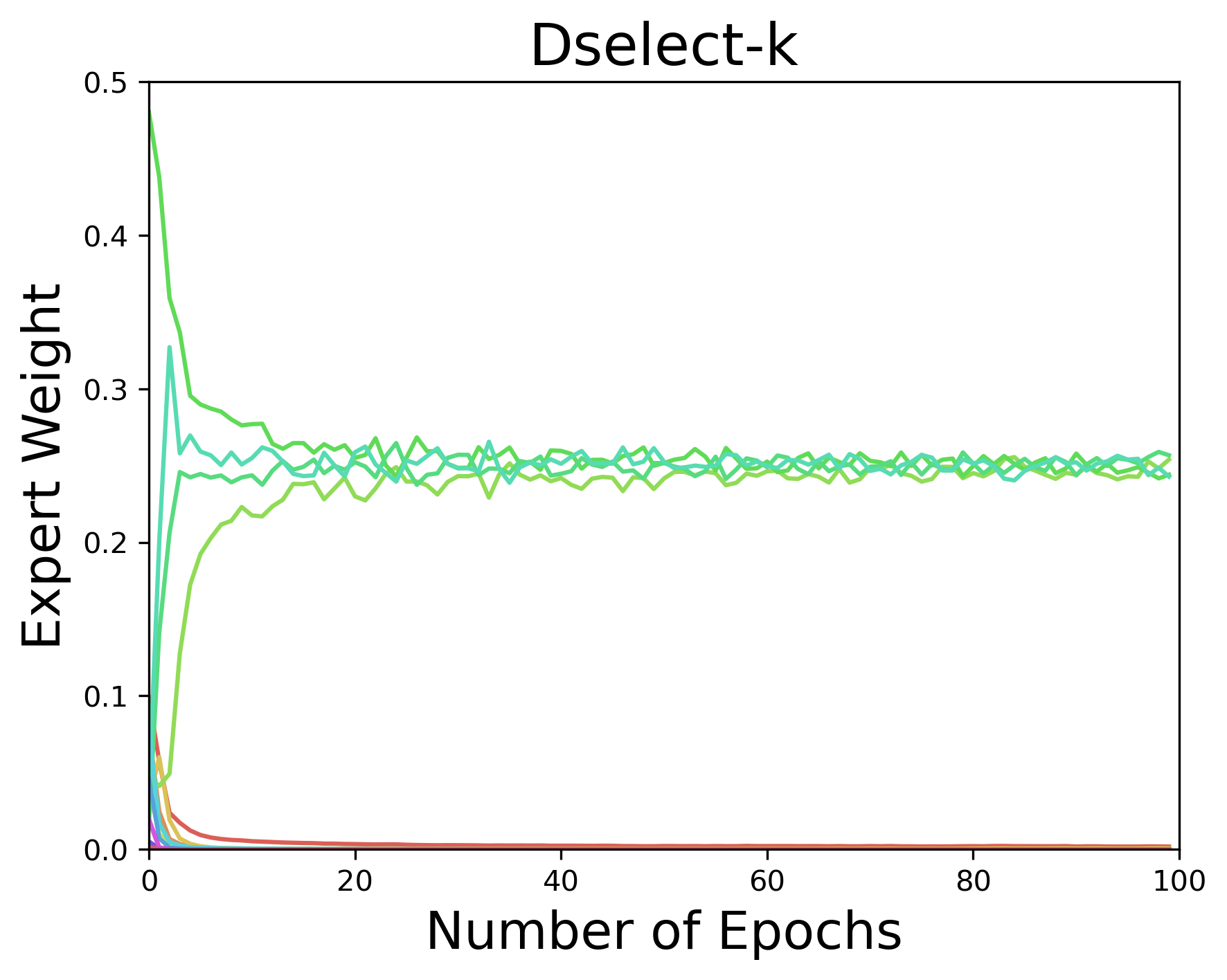}
    \caption{Expert weights output by Top-k (left) and DSelect-k (right) during training on synthetic data generated from a MoE, under static gating. Each color represents a separate expert. Here DSelect-k recovers the true experts used by the data-generating model, whereas Top-k does not recover and exhibits oscillatory behavior. See Appendix \ref{sec:visualization_appendix} for details on the data and setup.}
    \label{fig:dselect-k_viz_main}
\end{figure}

\textbf{Top-k Gate:} This is a sparse variant of the softmax gate that returns a probability vector with only k nonzero entries \citep{shazeer2017outrageously}. The Top-k gate is defined by $\sigma(\textsl{KeepTopK}(A x + b))$, where for any vector $v$, $\textsl{KeepTopK}(v)_i := v_i$ if $v_i$ is in the top k elements of $v$, and $\textsl{KeepTopK}(v)_i := - \infty$ otherwise\footnote{To balance the load across experts, \cite{shazeer2017outrageously} add noise and additional regularizers to the model.}. This gate is conceptually appealing since it allows for direct control over the number of experts to select and is trained using SGD.
Moreover, the Top-k gate supports \textsl{conditional training}: in backpropagation, for each input example, only the gradients of the loss w.r.t. top k elements need to be computed. With a careful implementation, conditional training can lead to computational savings. However, the Top-k gate is not continuous, which implies that the gradient does not exist at certain inputs. This can be problematic when training is done using gradient-based  methods. To gain more insight, in Figure  \ref{fig:dselect-k_viz_main} (left), we plot the expert weights chosen by the Top-k gate during training with SGD. The results indicate an oscillatory behavior in the output of the Top-k gate, which can be attributed to its discontinuous nature: a small change in the input can lead to ``jumps'' in the output.

\textbf{Comparison with DSelect-k:} We develop DSelect-k in Section \ref{sec:our_gate}. Here we present a high-level comparison between DSelect-k and Top-k. Similar to Top-k, DSelect-k can select $k$ out of the $n$ experts and can be trained using gradient-based optimization methods. A major advantage of DSelect-k over Top-k is that it is continuously differentiable, which leads to more stable selection of experts during training---see Figure  \ref{fig:dselect-k_viz_main} (right). During inference, DSelect-k only needs to evaluate a subset of the experts, which can lead to computational savings. However, DSelect-k supports conditional training only partially. At the start of training, it uses all the available experts, so conditional training is not possible. As we discuss in Section \ref{sec:our_gate}, after a certain point during training, DSelect-k converges to a small subset of the experts, and then conditional training becomes possible. Our experiments indicate that DSelect-k can have a  significant edge over Top-k in terms of prediction and expert selection performance, so the full support for conditional training in Top-k seems to come at the expense of statistical performance.
\section{Differentiable and Sparse Gating} \label{sec:our_gate}
In this section, we develop DSelect-k, for both the static and per-example gating settings. First, we introduce the problem setup and notation. To simplify the presentation, we will develop the gate for a single supervised learning task, and we note that the same gate can be used in MTL models. We assume that the task has an input space $\mathcal{X} \subset \mathbb{R}^{p}$, an output space $\mathcal{Y}$, and an associated loss function $\ell: \mathcal{Y} \times \mathbb{R} \to \mathbb{R}$. We denote the set of $N$ training examples by $\mathcal{D} = \{ (x_i, y_i) \in \mathcal{X} \times \mathcal{Y}\}_{i=1}^{N}$. We consider a learning model defined by the MoE in Equation \eqref{eq:moe_output}. For simplicity, we assume that the experts are scalar-valued and belong to a class of continuous  functions $\mathcal{H}$. We assume that the number of experts $n = 2^m$ for some integer $m$---in Appendix \ref{sec:arbitrary_n}, we discuss how the gate can be extended to arbitrary $n$. For convenience, given a non-negative integer $i$, we denote the set $\{1, 2, \dots, i\}$ by $[i]$.

In Section \ref{sec:static_gating}, we develop DSelect-k for static gating setting. Then, in Section \ref{sec:per-example_gating}, we generalize it to the per-example setting.

\vspace{-0.2cm}
\subsection{DSelect-k for Static Gating}
\label{sec:static_gating}
\vspace{-0.2cm}
Our goal here is to develop a static gate that selects a convex combination of at most $k$ out of the $n$ experts. The output of the gate can be thought of as a probability vector $w$ with at most $k$ nonzero entries, where $w_i$ is the weight assigned to the expert $f_i$. A natural way to minimize the empirical risk of the MoE model is by solving the following problem:
\vspace{0cm}
\begin{subequations} \label{eq:constrained_example}
\begin{align}
    \min_{f_1, \dots f_n, w} ~~~~ & \frac{1}{N} \sum_{(x, y) \in \mathcal{D}} \ell\Big(y, \sum_{i = 1}^{n} f_i(x) w_i \Big) \\
    \mathrm{s.t.} ~~~~~~~~ & \| w \|_0 \leq k \label{eq:cardinality_constraint} \\
    & \sum_{i=1}^{n} w_i = 1 , ~ w \geq 0. \label{eq:simplex_constraints}
\end{align} 
\end{subequations}
In the above, the $L_{0}$ norm of $w$, $\| w \|_0 $, is equal to the number of nonzero entries in $w$. Thus, the cardinality constraint \eqref{eq:cardinality_constraint} ensures that the gate selects at most $k$ experts. Problem \eqref{eq:constrained_example} is a combinatorial optimization problem that is not amenable to SGD due to the cardinality constraint \eqref{eq:cardinality_constraint} and the simplex constraints in \eqref{eq:simplex_constraints}. In what follows of this section, we first transform Problem \eqref{eq:constrained_example} into an equivalent unconstrained optimization problem, based on a binary encoding scheme. However, the unconstrained problem cannot be directly handled using SGD due to the presence of binary variables. Thus, in a second transformation, we smooth the binary variables, which leads to an optimization problem that is amenable to SGD.


\textbf{Road map: } In Section \ref{sec:single_expert}, we introduce the \textsl{single expert selector}: a construct for  choosing 1 out of $n$ experts by using binary encoding. In Section \ref{sec:combinatorial_gating}, we leverage the single expert selector to transform Problem \eqref{eq:constrained_example} into an unconstrained one. Then, in Section \ref{sec:smooth_gating}, we smooth the unconstrained problem and discuss how SGD can be applied.
\vspace{-0.2cm}
\subsubsection{Single Expert Selection using Binary Encoding} \label{sec:single_expert}
The single expert selector (selector, for short) is a fundamental construct that we will later use to convert Problem \eqref{eq:constrained_example} to an unconstrained optimization problem. At a high-level, the single expert selector chooses the index of 1 out of the $n$ experts and returns a one-hot encoding of the choice. For example, in the case of $4$ experts, the selector can choose the first expert by returning the binary vector $[1 ~ 0 ~ 0 ~ 0]^T$. Generally, the selector can choose any of the experts, and its choice is determined by a set of binary encoding variables, as we will describe next.

The selector is parameterized by $m$ (recall that $m = \log_2{n}$)  binary variables, $z_1, z_2, \dots, z_m$, where we view these variables collectively as a binary number: $z_m z_{m-1} \dots z_1$. The integer represented by the latter binary number determines which expert to select. More formally, let $l$ be the integer represented by the binary number $z_m z_{m-1} \dots z_1$. The selector is a function ${r}: \mathbb{R}^{m} \to \{0,1\}^{n}$ which maps $z := [z_1, z_2, \dots, z_m]^T$ to a one-hot encoding of the integer $(l+1)$. For example, if all the $z_i$'s are $0$, then the selector returns a one-hot encoding of the integer $1$. Next, we define the selector $r(z)$. For easier exposition, we start with the special case of $4$ experts and then generalize to $n$ experts.


\textbf{Special case of $4$ experts:} In this case, the selector uses two binary variables $z_1$ and $z_2$. Let $l$ be the integer represented by the binary number $z_2 z_1$. Then, the selector should return a one-hot encoding of the integer $(l+1)$. To achieve this, we define the selector $r(z)$ as follows:
\begin{align} \label{eq:r_z_4}
r(z) = 
\begin{bmatrix}
\bar{z_1} \bar{z_2}, ~~
{z_1} \bar{z_2}, ~~
\bar{z_1} {z_2}, ~~
{z_1} {z_2} 
\end{bmatrix}^{T}
\end{align}
where $\bar{z_i} := 1 - z_i$. By construction, exactly one entry in $r(z)$ is 1 (specifically, $r(z)_{l+1} = 1$) and the rest of the entries are zero. For example, if $z_1 = z_2 = 0$, then $r(z)_1 = 1$ and $r(z)_i = 0$, $i \in \{2,3,4\}$.

\textbf{General case of $n$ experts:} Here we generalize the selector $r(z)$ to the case of $n$ experts. To aid in the presentation, we make the following definition. For any non-negative integer $l$, we define $\mathcal{B}(l)$ as the set of indices of the nonzero entries in the binary representation of $l$ (where we assume that the least significant bit is indexed by $1$). For example, $\mathcal{B}(0) = \emptyset$, $\mathcal{B}(1) = \{1\}$,  $\mathcal{B}(2) = \{2\}$, and $\mathcal{B}(3) = \{1, 2\}$. For every $i \in [n]$, we define the $i$-th entry of $r(z)$ as follows:
\begin{align} \label{eq:gate_single}
    {r(z)}_i = \prod_{j \in \mathcal{B}(i-1)} (z_j)   \prod_{j \in [m] \setminus \mathcal{B}(i-1)} (1 - z_j) 
\end{align}
In the above, $r(z)_i$ is a product of $m$ binary variables, which is equal to $1$ iff the integer $(i-1)$ is represented by the binary number $z_m z_{m-1} \dots z_1$. Therefore, $r(z)$ returns a one-hot encoding of the index of the selected expert. Note that when $n=4$, definitions   \eqref{eq:r_z_4} and \eqref{eq:gate_single} are equivalent.
\vspace{-0.1cm}
\subsubsection{Multiple Expert Selection via Unconstrained Minimization} \label{sec:combinatorial_gating}
\vspace{-0.1cm}
In this section, we develop a combinatorial gate that allows for transforming Problem \eqref{eq:constrained_example} into an  unconstrained optimization problem. We design this gate by creating $k$ instances of the single expert selector $r(.)$, and then taking a convex combination of these $k$ instances. More formally, for every $i \in [k]$, let $z^{(i)} \in \{0,1\}^{m}$ be a (learnable) binary  vector, so that the output of the $i$-th instance of the selector is $r(z^{(i)})$. Let $Z$ be a $k \times m$ matrix whose $i$-th row is $z^{(i)}$. Moreover, let $\alpha \in \mathbb{R}^{k}$ be a vector of learnable parameters. We define the \textsl{combinatorial gate} $q$ as follows:
\begin{align*}
    q(\alpha, Z) = \sum_{i=1}^k {\sigma}(\alpha)_i  {r}(z^{(i)}),
\end{align*}
where we recall that $\sigma(.)$ is the softmax function. Since for every $i \in [k]$,  $r(z^{(i)})$ is a one-hot vector, we have $\| q(\alpha, Z) \|_0 \leq k$. Moreover, since the weights of the selectors are obtained using a softmax, we have $q(\alpha, Z) \geq 0$ and $\sum_{i=1}^{n} q(\alpha, Z)_i = 1$. Thus, $q(\alpha, Z)$ has the same interpretation of $w$ in Problem \eqref{eq:constrained_example}, without requiring any constraints. Therefore, we propose replacing $w$ in the objective of Problem \eqref{eq:constrained_example} with $q(\alpha, Z)$ and removing all the constraints. This replacement leads to an equivalent unconstrained optimization problem, as we state in the next proposition.
\begin{proposition} \label{prop:equivalence}
Problem \eqref{eq:constrained_example} is equivalent\footnote{Equivalent means that the two problems have the same optimal objective, and given an optimal solution for one problem, we can construct an optimal solution for the other.} 
to:
\begin{align} \label{eq:unconstrained_example}
\begin{split}
   \min_{f_1, \dots f_n, \alpha, Z} ~~~~ & \frac{1}{N} \sum_{(x, y) \in \mathcal{D}} \ell\Big(y, \sum_{i = 1}^{n} f_i(x)  q(\alpha, Z)_i  \Big) \\
    & z^{(i)} \in \{ 0, 1 \}^{m}, ~ i \in [k]
\end{split}
\end{align}
\end{proposition}
\vspace{-0.1cm}
The proof of Proposition \ref{prop:equivalence} is in Appendix \ref{appendix:proofs}. Unlike Problem \eqref{eq:constrained_example}, Problem \eqref{eq:unconstrained_example} does not involve any constraints, aside from requiring binary variables. However, these binary variables cannot be directly handled using first-order methods. Next, we discuss how to smooth the binary variables in order to obtain a continuous relaxation of Problem \eqref{eq:unconstrained_example}.

\subsubsection{Smooth Gating} \label{sec:smooth_gating}
\vspace{-0.1cm}
In this section, we present a procedure to smooth the binary variables in Problem \eqref{eq:unconstrained_example} and discuss how the resulting problem can be optimized using first-order methods. The procedure relies on the \textsl{smooth-step} function, which we define next.

\textbf{Smooth-step Function: } This is a continuously differentiable and S-shaped function, similar in shape to the logistic function. However, unlike the logistic function, the smooth-step function can output 0 and 1 exactly for sufficiently large magnitudes of the input. The smooth-step and logistic functions are depicted in Appendix \ref{sec:smooth_step}. More formally, given a non-negative scaling parameter $\gamma$, the smooth-step function, $S: \mathbb{R} \to \mathbb{R}$, is a cubic piecewise polynomial defined as follows:
$$
    S(t) = 
        \begin{cases}
            0 & \text{ if } t \leq -\gamma/2 \\
            -\frac{2}{\gamma^{3}}t^3 + \frac{3}{2\gamma}t + \frac{1}{2} & \text{ if } -\gamma/2 \leq t \leq \gamma/2 \\
            1 & \text{ if } t \geq \gamma/2
        \end{cases}
$$
The parameter $\gamma$ controls the width of the fractional region (i.e., the region where the function is strictly between 0 and 1). Note that $S(t)$ is continuously differentiable at all points---this follows since at the boundary points $\pm \gamma/2$, we have: $S'(-\gamma/2) = S'(\gamma/2) = 0$. This function has been recently used for conditional computation in soft trees \cite{hazimeh2020tree} and is popular in the  computer graphics literature  \citep{ebert2003texturing,rost2009opengl}.


\textbf{Smoothing: } We obtain DSelect-k from the combinatorial gate $q(\alpha, Z)$ by (i) relaxing every binary variable in $Z$ to be continuous in the range $(-\infty, +\infty)$, i.e., $Z \in \mathbb{R}^{k \times m}$, and (ii) applying the smooth-step function to $Z$ element-wise. Formally, DSelect-k is a function $\tilde{q}$ defined as follows:
\begin{align} \label{eq:smoothed_router}
        \tilde{q}(\alpha, Z) := q(\alpha, S(Z)) = \sum_{i=1}^k {\sigma}(\alpha)_i  {r}\big(S(z^{(i)})\big),
\end{align}
where the matrix $S(Z)$ is obtained by applying $S(\cdot)$ to $Z$ element-wise. Note that $\tilde{q}(\alpha, Z)$ is continuously differentiable so it is amenable to first-order methods. If $S(Z)$ is binary, then $ \tilde{q}(\alpha, Z)$ selects at most $k$ experts (this holds since $\tilde{q}(\alpha, Z) = q(\alpha, S(Z))$, and from Section \ref{sec:combinatorial_gating}, $q$ selects at most $k$ experts when its encoding matrix is binary). However, when $S(Z)$ has any non-binary entries, then more than $k$ experts can be potentially selected, meaning that the cardinality  constraint will not be respected. In what follows, we discuss how the gate can be optimized using first-order methods, while ensuring that $S(Z)$ converges to a binary matrix so that the cardinality constraint is enforced.

We propose using $\tilde{q}(\alpha, Z)$ in MoE, which leads to the following optimization problem:
\begin{align} \label{eq:smooth_optimization_no_entropy}
\begin{split}
    \min_{f_1, \dots f_n, \alpha, Z} ~~~~ & \frac{1}{N} \sum_{(x, y) \in \mathcal{D}} \ell\Big(y, \sum_{i = 1}^{n} f_i(x)  \tilde{q}(\alpha, Z)_i  \Big).
\end{split}
\end{align}
Problem \eqref{eq:smooth_optimization_no_entropy} can be viewed as a continuous relaxation of Problem \eqref{eq:unconstrained_example}. If the experts are differentiable, then the objective of Problem \eqref{eq:smooth_optimization_no_entropy} is differentiable. Thus, we propose optimizing MoE end-to-end using first-order methods. \textcolor{black}{We note that $\tilde{q}(\alpha, Z)$ uses $(k + k \log{n})$ learnable parameters. In contrast, the Top-k and softmax gates (discussed in Section \ref{sec:MoE}) use $n$ parameters. Thus, for relatively small $k$, our proposal uses a smaller number of parameters.} Next, we discuss how  DSelect-k's parameters should be initialized in order to ensure that it is trainable.


\textbf{Initialization: } By the definition of the smooth-step function, if $S(Z_{ij})$ is binary then $S'(Z_{ij}) = 0$, and consequently $\frac{\partial \ell}{\partial Z_{ij}} = 0$. This implies that, during optimization, if $S(Z_{ij})$ becomes binary, the variable $Z_{ij}$ will not be updated in any subsequent iteration. Thus, we have to be careful about the initialization of $Z$. For example, if $Z$ is initialized so that $S(Z)$ is a binary matrix then the gate will not be trained. To ensure that the gate is trainable, we initialize each $Z_{ij}$ so that $0 < S(Z_{ij}) < 1$. This way, the $Z_{ij}$'s can have  nonzero gradients at the start of optimization. 

\textbf{Accelerating Convergence to Binary Solutions: }
Recall that we need $S(Z)$ to converge to a binary matrix, in order for the gate $\tilde{q}$ to respect the cardinality constraint (i.e., to select at most $k$ experts). Empirically, we observe that if the optimizer runs for a sufficiently large number of iterations, then $S(Z)$ typically converges to a binary matrix. However, early stopping of the optimizer can be desired in practice for computational and statistical considerations, and this can prevent $S(Z)$ from converging. To encourage faster convergence towards a binary $S(Z)$, we will add an entropy regularizer to Problem  \eqref{eq:smooth_optimization_no_entropy}. The following proposition is needed before we introduce the regularizer.
\begin{proposition} \label{prop:simplex}
For any $z \in \mathbb{R}^{m}$, $\alpha \in \mathbb{R}^{k}$, and $Z \in \mathbb{R}^{k \times m}$, ${r}(S(z))$ and $\tilde{q}(\alpha, Z)$ belong to the probability simplex.
\end{proposition}
The proof of the proposition is in Appendix \ref{appendix:proofs}. Proposition \ref{prop:simplex} implies that, during training, the output of each single expert selector used by $\tilde{q}(\alpha, Z)$, i.e., $r(S(z^{(i)}))$ for $i \in [k]$, belongs to the probability simplex. Note that the entropy of each $r(S(z^{(i)}))$ is minimized by any one-hot encoded vector. Thus, for each $r(S(z^{(i)}))$, we add an entropy regularization term that encourages convergence towards one-hot encoded vectors;  equivalently, this encourages convergence towards a binary $S(Z)$. Specifically, we solve the following regularized variant of Problem \eqref{eq:smooth_optimization_no_entropy}:
\begin{align*}
\begin{split}
       \min_{f_1, \dots f_n, \alpha, Z} ~~~~ &  \sum_{(x, y) \in \mathcal{D}} \frac{1}{N} \ell\Big(y, \sum_{i = 1}^{n} f_i(x)  \tilde{q}(\alpha, Z)_i  \Big) + \lambda \Omega(Z)
\end{split}
\end{align*}
where $\Omega(Z) := \sum_{i=1}^{k} h \big(r(S(z^{(i)})) \big)$ and $h(.)$ is the entropy function. The hyperparameter $\lambda$ is non-negative and controls how fast each selector converges to a one-hot encoding. In our experiments, we tune over a range of $\lambda$ values. When selecting the best hyperparameters from tuning, we disregard any $\lambda$ whose corresponding solution does not have a binary $S(Z)$. In Appendix \ref{sec:convergence_appendix}, we report the number of training steps required for $S(Z)$ to converge to a binary matrix, on several real datasets. Other alternatives to ensure that $S(Z)$ converges to a binary matrix are also possible. One alternative is to regularize the entropy of each entry in $S(Z)$ separately. Another alternative is to anneal the parameter $\gamma$ of the smooth-step function towards zero. 

\textcolor{black}{\textbf{Softmax-based Alternative to Binary Encoding:} Recall that our proposed selectors in \eqref{eq:smoothed_router}, i.e., ${r}\big(S(z^{(i)})\big)$, $i \in [k]$, learn one-hot vectors \textsl{exactly} (by using binary encoding). 
One practical alternative for learning a one-hot vector is by using a softmax function with temperature annealing. Theoretically, this alternative cannot return a one-hot vector, but after training, the softmax output can be transformed to a one-hot vector using a heuristic (e.g., by taking an argmax). In Appendix \ref{sec:synthetic}, we perform an ablation study  in which we replace the selectors in DSelect-k with softmax functions (along with temperature annealing or entropy regularization).}



\subsection{DSelect-k for Per-example Gating} \label{sec:per-example_gating}
In this section, we generalize the static version of DSelect-k, $\tilde{q}(\alpha, Z)$, to the per-example gating setting. The key idea is to make the gate's parameters $\alpha$ and $Z$ functions of the input, so that the gate can make decisions on a per-example basis. Note that many functional forms are possible for these parameters. For simplicity and based on our experiments, we choose to make $\alpha$ and $Z$ linear functions of the input example. More formally, let $G \in \mathbb{R}^{k \times p}$, $W^{(i)} \in \mathbb{R}^{m \times p}$, $i \in [k]$, be a set of learnable parameters. Given an input example $x \in \mathbb{R}^{p}$, we set $\alpha = G x$ and $z^{(i)} = W^{(i)} x$ in $\tilde{q}(\alpha, Z)$ (to simplify the presentation, we do not include bias terms). Thus, the per-example version of DSelect-k is a function $v$ defined as follows:
$$
    v(G, W, x) = \sum_{i=1}^k {\sigma}(G x)_i  r\big(S( W^{(i)} x)\big).
$$
In the above, the term $r\big(S( W^{(i)} x)\big)$ represents the $i$-th single expert selector, whose output depends on the example $x$; thus different examples are free to select different experts. The term ${\sigma}(G x)_i$ determines the input-dependent weight assigned to the $i$-th selector. The gate  $v(G, W, x)$ is continuously differentiable in the parameters $G$ and $W$, so we propose optimizing it using first-order methods. Similar to the case of static gating, if $S( W^{(i)} x)$ is binary for all $i \in [k]$, then each $r\big(S( W^{(i)} x)\big)$ will select exactly one expert, and the example $x$ will be assigned to at most $k$ experts. 

To encourage $S( W^{(i)} x)$, $i \in [k]$ to become binary, we introduce an entropy regularizer, similar in essence to that in static gating. However, unlike static gating, the regularizer here should be on a per-example basis, so that each example respects the cardinality constraint. By Proposition \ref{prop:simplex}, for any $i \in [k]$, $r\big(S( W^{(i)} x)\big)$ belongs to the probability simplex. Thus, for each example $x$ in the training data, we introduce a  regularization term of the form: $ \Omega(W, x) := \sum_{i \in [k]} h\Big(r\big(S( W^{(i)} x)\big) \Big)$, and minimize the following objective function:
\begin{align*}
    \sum_{(x, y) \in \mathcal{D}}  \Bigg( \frac{1}{N} \ell \Big( y, \sum_{i = 1}^{n} f_i(x)  v(G, W, x)_i  \Big) + \lambda \Omega(W, x) \Bigg),
\end{align*}
where $\lambda$ is a non-negative hyperparameter. Similar to the case of static gating, we tune over a range of $\lambda$ values, and we only consider the  choices of $\lambda$ that force the average number of selected experts per example to be less than or equal to $k$. If the application requires that the cardinality  constraint be satisfied strictly for every example (not only on average), then annealing $\gamma$ in the smooth-step function towards zero enforces this.

\vspace{-0.2cm}
\section{Experiments} \label{sec:experiments}
\vspace{-0.2cm}
We study the performance of DSelect-k in the context of MTL and compare with state-of-the-art gates and baselines. In the rest of this section, we present experiments on the following real MTL datasets: MovieLens, Multi-MNIST, Multi-Fashion MNIST, \textcolor{black}{and on a real-world,  large-scale recommender system.} \textcolor{black}{Moreover, in Appendix \ref{sec:appendix_experiments}, we present an additional  experiment on synthetic data (with up to $128$ tasks), in which we study statistical  performance and perform ablation studies.}


\textbf{Competing Methods: } We focus on a multi-gate MoE, and study the DSelect-k and Top-k gates in both the static and per-example gating settings. \textcolor{black}{For static gating, we also consider a Gumbel-softmax based gate \citep{sun2019adashare}--unlike DSelect-k this gate cannot control the sparsity level explicitly (see the supplementary for details)}. In addition, we consider two MTL baselines. The first  baseline is a MoE with a softmax gate (which uses all the available experts). The second is a \textsl{shared bottom} model \citep{caruana1997multitask}, where all tasks share the same bottom layers, which are in turn connected to task-specific neural nets.


\textbf{Experimental Setup: } All competing models were implemented in TensorFlow 2. We used Adam \citep{kingma2014adam} and Adagrad \citep{duchi2011adaptive} for optimization, and we tuned the key hyperparameters using random grid search (with an average of $5$ trials per grid point). Full details on the setup are in Appendix \ref{sec:experimental_details}.  
\vspace{-0.2cm}
\subsection{MovieLens}
\vspace{-0.2cm}
\textbf{Dataset:} MovieLens \citep{harper2015movielens} is a movie recommendation dataset containing records for $4{,}000$ movies and $6{,}000$ users. Following \cite{wang2020small}, for every user-movie pair, we construct two tasks. Task 1 is a binary classification problem for predicting whether the user will watch a particular movie. Task 2 is a regression problem to predict the user's rating (in $\{1,2,\dots, 5\}$) for a given movie. We use $1.6$ million examples for training and $200,000$ for each of the validation and testing sets. 

\textbf{Experimental Details:} We use the cross-entropy and squared error losses for tasks 1 and 2, respectively. We optimize a weighted average of the two losses, i.e., the final loss function is $\alpha (\text{Loss of Task 1}) + (1-\alpha)  (\text{Loss of Task 2}) $, and we report the results for $\alpha \in \{0.1, 0.5, 0.9 \}$. The same loss function is also used for tuning and testing. The architecture consists of a multi-gate MoE with $8$ experts, where each of the experts and the task-specific networks is composed of ReLU-activated dense layers. For each $\alpha$, we tune over the optimization and gate-specific hyperparameters, including the number of experts to select (i.e., k in DSelect-k and Top-k). After tuning, we train each model for $100$ repetitions (using random initialization) and report the averaged results. For full details, see Appendix \ref{sec:movie_lens_extra_details}.

\textbf{Results:} In Table \ref{table:movielens}, we report the test loss and the average number of selected experts. 
The results indicate that for all values of $\alpha$, either one of our DSelect-k gates (static or per-example) outperforms the competing methods, in terms of both the test loss and the number of selected experts. \textcolor{black}{In the static gating setting, there does not seem to be a clear winner among the three competing methods (Top-k, DSelect-k, and Gumbel Softmax), but we note that DSelect-k outperforms both Top-k and Gumbel Softmax for two out of the three choices of $\alpha$.} Notably, the softmax MoE is uniformly outperformed by the DSelect-k and Top-k gates, so sparsity in gating seems to be beneficial on this dataset. \textcolor{black}{Our hypothesis is that softmax MoE is overfitting and the sparse gating methods are mitigating this issue.} 
\textcolor{black}{In Table \ref{table:movie_lens_extra_table} in the appendix, we additionally report the individual task metrics (loss and accuracy).}

\begin{table}[htbp]
\centering
\caption{\small{Test loss (with standard error) and average number of selected experts on  MovieLens. The parameter $\alpha$ is the weight of Task 1's loss (see text for details). The test loss is  multiplied by $10^4$.}}
\label{table:movielens}
\resizebox{0.9\columnwidth}{!}{
\begin{tabular}{c|l|cc|cc|cc|}
\cline{3-8}
\multicolumn{1}{l}{}                                                                                & \multirow{2}{*}{} & \multicolumn{2}{c|}{$\alpha = 0.1$} & \multicolumn{2}{c|}{$\alpha = 0.5$} & \multicolumn{2}{c|}{$\alpha = 0.9$} \\
\multicolumn{1}{l}{}                                                                                &                         & Loss                & Experts       & Loss                & Experts       & Loss                  & Experts     \\ \hline
\multicolumn{1}{|c|}{\multirow{3}{*}{\begin{tabular}[c]{@{}c@{}}Static\end{tabular}}}      & DSelect-k           & $4015 \pm 5$        & 2.7           & $\bm{3804} \pm 3$        & \bm{$1.5$}           & ${3690} \pm 2$          & \bm{$1.3$}        \\
\multicolumn{1}{|c|}{}                                                                               & Top-k          & $\bm{4012} \pm 4$        & \bm{$2.0$}           & $3818 \pm 2$        & 2.0           & $3693 \pm 6$          & 2.0         \\
\multicolumn{1}{|c|}{}                                                                               & Gumbel Softmax & $4171 \pm 3$        & 2.7           & $3898 \pm 2$        & 2.6           & $\bm{3688} \pm 4$          & 3.6         \\ \hline
\multicolumn{1}{|c|}{\multirow{2}{*}{\begin{tabular}[c]{@{}c@{}}Per-example\end{tabular}}} & DSelect-k      & $\bm{4006} \pm 6$        & \bm{$1.5$}           & $\bm{3823} \pm 3$        & \bm{$1.2$}           & $\bm{3679} \pm 2$          & $\bm{1.1}$         \\
\multicolumn{1}{|c|}{}                                                                               & Top-k     & $4027 \pm 8$        & 2.0           & $3841 \pm 4$        & 2.0           & $3741 \pm 3$          & 2.0         \\ \hline
\multicolumn{1}{|c|}{\multirow{2}{*}{Baselines}}                                                     & Softmax MoE                 & $4090 \pm 1$        & 8.0           & $3960 \pm 3$        & 8.0           & $3847 \pm 10$      & 8.0         \\
\multicolumn{1}{|c|}{}                                                                               & Shared Bottom           & $4037 \pm 2$        & -             & $3868 \pm 2$        & -             & $3687 \pm 1$          & -           \\ \hline
\end{tabular}
}
\vspace{-0.2cm}
\end{table}
\vspace{-0.5cm}
\subsection{Multi-MNIST and Multi-Fashion MNIST}
\vspace{-0.2cm}
\textbf{Datasets:} We consider two image classification datasets:  Multi-MNIST  and Multi-Fashion \citep{sabour2017dynamic}, which are multi-task variants of the MNIST \citep{lecun2010mnist} and Fashion MNIST \citep{xiao2017fashion} datasets. We construct the Multi-MNIST dataset similar to \cite{sabour2017dynamic}: uniformly sample two images from MNIST and overlay them on top of each other, and (ii) shift one digit towards the top-left corner and the other digit towards the bottom-right corner (by $4$ pixels in each direction). This procedure leads to $36 \times 36$ images with some overlap between the digits. The Multi-Fashion is constructed in a similar way by overlaying images from the Fashion MNIST dataset. For each dataset, we consider two classification tasks: Task 1 is to classify the top-left item and Task 2 is to classify the bottom-right item. We use $100{,}000$ examples for training, and $20{,}000$ examples for each of the validation and testing sets.

\textbf{Experimental Details: } We use cross-entropy loss for each task and optimize the sum of the losses\footnote{Due to the symmetry in the problem, assigning the two tasks equal weights is a reasonable choice.}. The model is a multi-gate MoE with $8$ experts, where each expert is a convolutional neural network and each task-specific network is composed of a number of dense layers. We tune the optimization and gate-specific hyperparameters, including the number of experts to select, and use the average of the task accuracies as the tuning metric. After tuning, we train each model for $100$ repetitions (using random initialization) and report the averaged results. For full details, see Appendix \ref{sec:multi_mnist_extra_details}.

\textbf{Results: } In Table \ref{table:mnist}, we report the test accuracy and the number of selected experts for the Multi-MNIST and  Multi-Fashion  datasets. On Multi-MNIST, DSelect-k (static) outperforms Top-k and Gumbel Softmax, in terms of both task accuracies and number of selected experts. For example, it achieves over $1\%$ improvement in Task 2's accuracy compared to Top-k (static). DSelect-k (static) comes close to the performance of the Softmax MoE, but uses less experts ($1.7$ vs. $8$ experts). Here DSelect-k (per-example) does not offer improvement over the static variant (unlike the MovieLens dataset). On Multi-Fashion, we again see that DSelect-k (static) performs best in terms of accuracy. 

\begin{table*}[htbp]
\centering
\caption{\small{Test accuracy (with standard error) and number of selected experts on Multi-MNIST/Fashion.}}
\label{table:mnist}
\resizebox{0.94\columnwidth}{!}{
\begin{tabular}{cl|ccc|ccc|}
\cline{3-8}
\multicolumn{1}{l}{}                                                                                 &                & \multicolumn{3}{c|}{Multi-MNIST}              & \multicolumn{3}{c|}{Multi-Fashion MNIST}      \\
\multicolumn{1}{l}{}                                                                                 &                & Accuracy 1       & Accuracy 2       & Experts & Accuracy 1       & Accuracy 2       & Experts \\ \hline
\multicolumn{1}{|c|}{\multirow{3}{*}{\begin{tabular}[c]{@{}c@{}}Static\end{tabular}}}      & DSelect-k           & $\bm{92.56} \pm 0.03$ & $\bm{90.98} \pm 0.04$ & $\bm{1.7}$     & $\bm{83.78} \pm 0.05$ & $\bm{83.34} \pm 0.05$ & ${1.8}$     \\
\multicolumn{1}{|c|}{}                                                                               & Top-k          & $91.93 \pm 0.06$ & $90.03 \pm 0.08$ & 4       & $83.44 \pm 0.07$ & $82.66 \pm 0.08$ & 4       \\
\multicolumn{1}{|c|}{}                                                                               & Gumbel Softmax & $92.3 \pm 0.05$  & $90.63 \pm 0.05$ & 1.8     & $83.66 \pm 0.07$ & $83.28 \pm 0.05$ & $\bm{1.5}$     \\ \hline
\multicolumn{1}{|c|}{\multirow{2}{*}{\begin{tabular}[c]{@{}c@{}}Per-example\end{tabular}}} & DSelect-k           & $\bm{92.42}\pm 0.03$ & $\bm{90.7} \pm 0.03$  & $\bm{1.5}$     & $\bm{83.69} \pm 0.04$ & ${83.13} \pm 0.04$ & $\bm{1.5}$     \\
\multicolumn{1}{|c|}{}                                                                               & Top-k          & $92.27 \pm 0.03$ & $90.45 \pm 0.03$ & 4       & $83.66 \pm 0.04$ & $\bm{83.15} \pm 0.04$ & 4       \\ \hline
\multicolumn{1}{|c|}{\multirow{2}{*}{Baselines}}                                                     & Softmax MoE        & $92.61 \pm 0.03$ & $91.0 \pm 0.03$  & 8       & $83.48 \pm 0.04$ & $82.81 \pm 0.04$ & 8       \\
\multicolumn{1}{|c|}{}                                                                               & Shared Bottom  & $91.3 \pm 0.04$  & $89.47 \pm 0.04$ & -       & $82.05 \pm 0.05$ & $81.37 \pm 0.06$ & -       \\ \hline
\end{tabular}
}
\vspace{-0.2cm}
\end{table*}

\begin{figure}[htbp]
\resizebox{0.85\columnwidth}{!}{
\begin{floatrow}
\hspace{-1cm}
\capbtabbox{%
\begin{tabular}{|c|c|c|}
\hline
Tasks / Methods  & DSelect-k & Top-k \\
 \hline
E. Task 1 (AUC)  & \textbf{0.8103} $\pm$ 0.0002  & 0.7481 $\pm$ 0.0198  \\
 \hline
E. Task 2 (AUC)  & \textbf{0.8161} $\pm$ 0.0002  & 0.7624 $\pm$ 0.0169  \\
 \hline
E. Task 3 (RMSE) & \textbf{0.2874} $\pm$ 0.0002 & 0.3406 $\pm$ 0.0180  \\
 \hline
E. Task 4 (RMSE)   & \textbf{0.8781} $\pm$ 0.0014  & 1.1213 $\pm$ 0.0842  \\
 \hline
E. Task 5 (AUC)   & \textbf{0.7524} $\pm$ 0.0003  & 0.5966 $\pm$ 0.0529  \\
 \hline
S. Task 1 (AUC)   & \textbf{0.6133} $\pm$ 0.0066  & 0.5080 $\pm$ 0.0047  \\
 \hline
S. Task 2 (AUC)   & \textbf{0.8468} $\pm$ 0.0289  & 0.5981 $\pm$ 0.0616  \\
 \hline
S. Task 3 (AUC)   & \textbf{0.9259} $\pm$ 0.0008 & 0.6665 $\pm$ 0.0091  \\
 \hline
 \end{tabular}
}{%
  \caption{Average performance (AUC and RMSE) and standard error on a real-world recommender system with 8 tasks: ``E.'' and ``S.'' denote engagement and satisfaction tasks, respectively.\vspace{0.25cm}}%
  \label{tab:prod}
}
\ffigbox{%
    \includegraphics[scale=0.42]{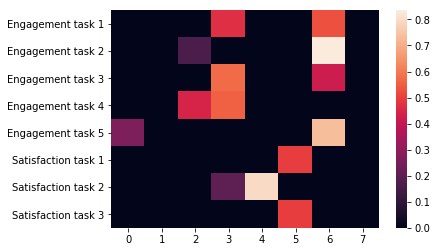}
}{%
  \caption{Expert weights of the DSelect-k gates on the recommender system.}%
  \label{fig:gating_vis}
}
\end{floatrow}
}
\end{figure}

\subsection{A Large-scale Recommender System}
\vspace{-0.1cm}
We study the performance of DSelect-k and Top-k in a real-word, large-scale content recommendation system. The system encompasses hundreds of millions of unique items and billions of users.

\textbf{Architecture and Dataset: }
The system consists of a candidate generator followed by a multi-task ranking model, and it adopts a framework similar to \cite{zhao2019recommending, tang2020progressive}. The ranking model makes predictions for 6 classification and 2 regression tasks. These can be classified into two categories: (i) engagement tasks (e.g., predicting user clicks, bad clicks, engagement time), and (ii) satisfaction tasks (e.g., predicting user satisfaction behaviors such as likes and dislikes). We construct the dataset from the system's user logs (which contain historical information about the user and labels for the $8$ tasks). The dataset consists of billions of examples (we do not report the exact number for confidentiality). We use a random 90/10 split for the training and evaluation sets.

\textbf{Experimental Details: } We use the cross-entropy and squared error losses for the classification and regression tasks, respectively. The ranking model is based on a multi-gate MoE, in which each task uses a separate static gate. The MoE uses $8$ experts, each composed of dense layers. For both the DSelect-k and Top-k based models, we tune the learning rate and the experts' architecture. Then, using the best hyperparameters, we train the final models for $5$ repetitions (using random initialization). For additional details, see Appendix \ref{sec:recommender_extra_details}.

\textbf{Results: }
In Table \ref{tab:prod}, we report the out-of-sample performance metrics for the $8$ tasks. The results indicate that DSelect-k outperforms Top-k on all tasks, with the improvements being most pronounced on the satisfaction tasks. In Figure \ref{fig:gating_vis}, we show a heatmap of the expert weights chosen by the DSelect-k gates. Notably, for DSelect-k, all engagement tasks share at least one expert, and two of the satisfaction tasks share the same expert.

\vspace{-0.2cm}
\section{Conclusion}
\vspace{-0.2cm}
We introduced DSelect-k: a continuously differentiable and sparse gate for MoE, which can be trained using first-order methods. Given a user-specified parameter $k$, the gate selects at most $k$ of the $n$ experts. Such direct control over the sparsity level is typically handled in the literature by adding a cardinality constraint to the optimization problem. One of the key ideas we introduced is a binary encoding scheme that allows for selecting $k$ experts, without requiring any constraints in the optimization problem. We studied the performance of DSelect-k in MTL settings, on both synthetic and real datasets. Our experiments indicate that DSelect-k can achieve significant improvements in prediction and expert selection, compared to state-of-the-art MoE gates and MTL baselines.

\textcolor{black}{\textbf{Societal Impact:} MoE models are used in various applications (as discussed in the introduction). DSelect-k can improve the interpretability and efficiency of MoE models, thus benefiting the underlying applications. We do not see direct, negative societal impacts from our proposal.}

\clearpage

\textbf{Acknowledgements:} 
The research was conducted while Hussein Hazimeh was at Google, and part of the writing was done during his time at MIT. At MIT, Hussein Hazimeh and Rahul Mazumder acknowledge research funding from the Office of Naval Research [Grant ONR-N000141812298].

\bibliographystyle{plainnat}
\bibliography{ref}

\begin{thebibliography}{41}
\providecommand{\natexlab}[1]{#1}
\providecommand{\url}[1]{\texttt{#1}}
\expandafter\ifx\csname urlstyle\endcsname\relax
  \providecommand{\doi}[1]{doi: #1}\else
  \providecommand{\doi}{doi: \begingroup \urlstyle{rm}\Url}\fi

\bibitem[Bengio et~al.(2015)Bengio, Bacon, Pineau, and
  Precup]{DBLP:journals/corr/BengioBPP15}
Emmanuel Bengio, Pierre{-}Luc Bacon, Joelle Pineau, and Doina Precup.
\newblock Conditional computation in neural networks for faster models.
\newblock \emph{CoRR}, abs/1511.06297, 2015.
\newblock URL \url{http://arxiv.org/abs/1511.06297}.

\bibitem[Bengio et~al.(2013)Bengio, L{\'e}onard, and
  Courville]{bengio2013estimating}
Yoshua Bengio, Nicholas L{\'e}onard, and Aaron Courville.
\newblock Estimating or propagating gradients through stochastic neurons for
  conditional computation.
\newblock \emph{arXiv preprint arXiv:1308.3432}, 2013.

\bibitem[Blondel et~al.(2019)Blondel, Martins, and
  Niculae]{blondel2019learning}
Mathieu Blondel, Andre Martins, and Vlad Niculae.
\newblock Learning classifiers with fenchel-young losses: Generalized
  entropies, margins, and algorithms.
\newblock In \emph{The 22nd International Conference on Artificial Intelligence
  and Statistics}, pages 606--615. PMLR, 2019.

\bibitem[Caruana(1997)]{caruana1997multitask}
Rich Caruana.
\newblock Multitask learning.
\newblock \emph{Machine learning}, 28\penalty0 (1):\penalty0 41--75, 1997.

\bibitem[Chen et~al.(2018)Chen, Song, Wainwright, and Jordan]{chen2018learning}
Jianbo Chen, Le~Song, Martin Wainwright, and Michael Jordan.
\newblock Learning to explain: An information-theoretic perspective on model
  interpretation.
\newblock In \emph{International Conference on Machine Learning}, pages
  883--892. PMLR, 2018.

\bibitem[Correia et~al.(2019)Correia, Niculae, and
  Martins]{correia2019adaptively}
Gon{\c{c}}alo~M Correia, Vlad Niculae, and Andr{\'e}~FT Martins.
\newblock Adaptively sparse transformers.
\newblock In \emph{Proceedings of the 2019 Conference on Empirical Methods in
  Natural Language Processing and the 9th International Joint Conference on
  Natural Language Processing (EMNLP-IJCNLP)}, pages 2174--2184, 2019.

\bibitem[Duchi et~al.(2011)Duchi, Hazan, and Singer]{duchi2011adaptive}
John Duchi, Elad Hazan, and Yoram Singer.
\newblock Adaptive subgradient methods for online learning and stochastic
  optimization.
\newblock \emph{Journal of machine learning research}, 12\penalty0 (7), 2011.

\bibitem[Ebert et~al.(2003)Ebert, Musgrave, Peachey, Perlin, and
  Worley]{ebert2003texturing}
David~S Ebert, F~Kenton Musgrave, Darwyn Peachey, Ken Perlin, and Steven
  Worley.
\newblock \emph{Texturing \& modeling: a procedural approach}.
\newblock Morgan Kaufmann, 2003.

\bibitem[Fedus et~al.(2021)Fedus, Zoph, and Shazeer]{fedus2021switch}
William Fedus, Barret Zoph, and Noam Shazeer.
\newblock Switch transformers: Scaling to trillion parameter models with simple
  and efficient sparsity.
\newblock \emph{arXiv preprint arXiv:2101.03961}, 2021.

\bibitem[Harper and Konstan(2015)]{harper2015movielens}
F~Maxwell Harper and Joseph~A Konstan.
\newblock The movielens datasets: History and context.
\newblock \emph{Acm transactions on interactive intelligent systems (tiis)},
  5\penalty0 (4):\penalty0 1--19, 2015.

\bibitem[Hazimeh et~al.(2020)Hazimeh, Ponomareva, Mol, Tan, and
  Mazumder]{hazimeh2020tree}
Hussein Hazimeh, Natalia Ponomareva, Petros Mol, Zhenyu Tan, and Rahul
  Mazumder.
\newblock The tree ensemble layer: Differentiability meets conditional
  computation.
\newblock In Hal~Daumé III and Aarti Singh, editors, \emph{Proceedings of the
  37th International Conference on Machine Learning}, volume 119 of
  \emph{Proceedings of Machine Learning Research}, pages 4138--4148, Virtual,
  13--18 Jul 2020. PMLR.

\bibitem[Ioannou et~al.(2016)Ioannou, Robertson, Zikic, Kontschieder, Shotton,
  Brown, and Criminisi]{ioannou2016decision}
Yani Ioannou, Duncan Robertson, Darko Zikic, Peter Kontschieder, Jamie Shotton,
  Matthew Brown, and Antonio Criminisi.
\newblock Decision forests, convolutional networks and the models in-between.
\newblock \emph{arXiv preprint arXiv:1603.01250}, 2016.

\bibitem[Jacobs(1997)]{jacobs1997bias}
Robert~A Jacobs.
\newblock Bias/variance analyses of mixtures-of-experts architectures.
\newblock \emph{Neural computation}, 9\penalty0 (2):\penalty0 369--383, 1997.

\bibitem[Jacobs et~al.(1991)Jacobs, Jordan, Nowlan, and
  Hinton]{jacobs1991adaptive}
Robert~A Jacobs, Michael~I Jordan, Steven~J Nowlan, and Geoffrey~E Hinton.
\newblock Adaptive mixtures of local experts.
\newblock \emph{Neural computation}, 3\penalty0 (1):\penalty0 79--87, 1991.

\bibitem[Jang et~al.(2016)Jang, Gu, and Poole]{jang2016categorical}
Eric Jang, Shixiang Gu, and Ben Poole.
\newblock Categorical reparameterization with gumbel-softmax.
\newblock \emph{arXiv preprint arXiv:1611.01144}, 2016.

\bibitem[Jiang and Tanner(1999)]{jiang1999identifiability}
Wenxin Jiang and Martin~A Tanner.
\newblock On the identifiability of mixtures-of-experts.
\newblock \emph{Neural Networks}, 12\penalty0 (9):\penalty0 1253--1258, 1999.

\bibitem[Jordan and Jacobs(1994)]{jordan1994hierarchical}
Michael~I Jordan and Robert~A Jacobs.
\newblock Hierarchical mixtures of experts and the em algorithm.
\newblock \emph{Neural computation}, 6\penalty0 (2):\penalty0 181--214, 1994.

\bibitem[Kingma and Ba(2015)]{kingma2014adam}
Diederik~P. Kingma and Jimmy Ba.
\newblock Adam: {A} method for stochastic optimization.
\newblock In Yoshua Bengio and Yann LeCun, editors, \emph{3rd International
  Conference on Learning Representations, {ICLR} 2015, San Diego, CA, USA, May
  7-9, 2015, Conference Track Proceedings}, 2015.

\bibitem[Kingma and Welling(2013)]{kingma2013auto}
Diederik~P Kingma and Max Welling.
\newblock Auto-encoding variational bayes.
\newblock \emph{arXiv preprint arXiv:1312.6114}, 2013.

\bibitem[LeCun et~al.(2010)LeCun, Cortes, and Burges]{lecun2010mnist}
Yann LeCun, Corinna Cortes, and CJ~Burges.
\newblock Mnist handwritten digit database.
\newblock \emph{ATT Labs [Online]. Available:
  http://yann.lecun.com/exdb/mnist}, 2, 2010.

\bibitem[Lepikhin et~al.(2020)Lepikhin, Lee, Xu, Chen, Firat, Huang, Krikun,
  Shazeer, and Chen]{lepikhin2020gshard}
Dmitry Lepikhin, HyoukJoong Lee, Yuanzhong Xu, Dehao Chen, Orhan Firat, Yanping
  Huang, Maxim Krikun, Noam Shazeer, and Zhifeng Chen.
\newblock Gshard: Scaling giant models with conditional computation and
  automatic sharding.
\newblock \emph{arXiv preprint arXiv:2006.16668}, 2020.

\bibitem[Ma et~al.(2018)Ma, Zhao, Yi, Chen, Hong, and Chi]{ma2018modeling}
Jiaqi Ma, Zhe Zhao, Xinyang Yi, Jilin Chen, Lichan Hong, and Ed~H Chi.
\newblock Modeling task relationships in multi-task learning with multi-gate
  mixture-of-experts.
\newblock In \emph{Proceedings of the 24th ACM SIGKDD International Conference
  on Knowledge Discovery \& Data Mining}, pages 1930--1939, 2018.

\bibitem[Maddison et~al.(2016)Maddison, Mnih, and Teh]{maddison2016concrete}
Chris~J Maddison, Andriy Mnih, and Yee~Whye Teh.
\newblock The concrete distribution: A continuous relaxation of discrete random
  variables.
\newblock \emph{arXiv preprint arXiv:1611.00712}, 2016.

\bibitem[Martins and Astudillo(2016)]{martins2016softmax}
Andre Martins and Ramon Astudillo.
\newblock From softmax to sparsemax: A sparse model of attention and
  multi-label classification.
\newblock In \emph{International Conference on Machine Learning}, pages
  1614--1623. PMLR, 2016.

\bibitem[Maziarz et~al.(2019)Maziarz, Kokiopoulou, Gesmundo, Sbaiz, Bartok, and
  Berent]{maziarz2019gumbel}
Krzysztof Maziarz, Efi Kokiopoulou, Andrea Gesmundo, Luciano Sbaiz, Gabor
  Bartok, and Jesse Berent.
\newblock Gumbel-matrix routing for flexible multi-task learning.
\newblock \emph{arXiv preprint arXiv:1910.04915}, 2019.

\bibitem[Paulus et~al.(2020)Paulus, Choi, Tarlow, Krause, and
  Maddison]{paulus2020gradient}
Max Paulus, Dami Choi, Daniel Tarlow, Andreas Krause, and Chris~J Maddison.
\newblock Gradient estimation with stochastic softmax tricks.
\newblock In H.~Larochelle, M.~Ranzato, R.~Hadsell, M.~F. Balcan, and H.~Lin,
  editors, \emph{Advances in Neural Information Processing Systems}, volume~33,
  pages 5691--5704. Curran Associates, Inc., 2020.
\newblock URL
  \url{https://proceedings.neurips.cc/paper/2020/file/3df80af53dce8435cf9ad6c3e7a403fd-Paper.pdf}.

\bibitem[Peters et~al.(2019)Peters, Niculae, and Martins]{peters2019sparse}
Ben Peters, Vlad Niculae, and Andr{\'e}~FT Martins.
\newblock Sparse sequence-to-sequence models.
\newblock In \emph{Proceedings of the 57th Annual Meeting of the Association
  for Computational Linguistics}, pages 1504--1519, 2019.

\bibitem[Ramachandran and Le(2018)]{ramachandran2018diversity}
Prajit Ramachandran and Quoc~V Le.
\newblock Diversity and depth in per-example routing models.
\newblock In \emph{International Conference on Learning Representations}, 2018.

\bibitem[Rosenbaum et~al.(2018)Rosenbaum, Klinger, and
  Riemer]{rosenbaum2018routing}
Clemens Rosenbaum, Tim Klinger, and Matthew Riemer.
\newblock Routing networks: Adaptive selection of non-linear functions for
  multi-task learning.
\newblock In \emph{International Conference on Learning Representations}, 2018.

\bibitem[Rost et~al.(2009)Rost, Licea-Kane, Ginsburg, Kessenich, Lichtenbelt,
  Malan, and Weiblen]{rost2009opengl}
Randi~J Rost, Bill Licea-Kane, Dan Ginsburg, John Kessenich, Barthold
  Lichtenbelt, Hugh Malan, and Mike Weiblen.
\newblock \emph{OpenGL shading language}.
\newblock Pearson Education, 2009.

\bibitem[Ruder(2017)]{ruder2017overview}
Sebastian Ruder.
\newblock An overview of multi-task learning in deep neural networks.
\newblock \emph{arXiv preprint arXiv:1706.05098}, 2017.

\bibitem[Sabour et~al.(2017)Sabour, Frosst, and Hinton]{sabour2017dynamic}
Sara Sabour, Nicholas Frosst, and Geoffrey~E Hinton.
\newblock Dynamic routing between capsules.
\newblock In \emph{Advances in neural information processing systems}, pages
  3856--3866, 2017.

\bibitem[Shazeer et~al.(2017)Shazeer, Mirhoseini, Maziarz, Davis, Le, Hinton,
  and Dean]{shazeer2017outrageously}
Noam Shazeer, Azalia Mirhoseini, Krzysztof Maziarz, Andy Davis, Quoc~V. Le,
  Geoffrey~E. Hinton, and Jeff Dean.
\newblock Outrageously large neural networks: The sparsely-gated
  mixture-of-experts layer.
\newblock In \emph{5th International Conference on Learning Representations,
  {ICLR} 2017, Toulon, France, April 24-26, 2017, Conference Track
  Proceedings}, 2017.

\bibitem[Sun et~al.(2020)Sun, Panda, Feris, and Saenko]{sun2019adashare}
Ximeng Sun, Rameswar Panda, Rogerio Feris, and Kate Saenko.
\newblock Adashare: Learning what to share for efficient deep multi-task
  learning.
\newblock In H.~Larochelle, M.~Ranzato, R.~Hadsell, M.~F. Balcan, and H.~Lin,
  editors, \emph{Advances in Neural Information Processing Systems}, volume~33,
  pages 8728--8740. Curran Associates, Inc., 2020.
\newblock URL
  \url{https://proceedings.neurips.cc/paper/2020/file/634841a6831464b64c072c8510c7f35c-Paper.pdf}.

\bibitem[Tang et~al.(2020)Tang, Liu, Zhao, and Gong]{tang2020progressive}
Hongyan Tang, Junning Liu, Ming Zhao, and Xudong Gong.
\newblock Progressive layered extraction (ple): A novel multi-task learning
  (mtl) model for personalized recommendations.
\newblock In \emph{Fourteenth ACM Conference on Recommender Systems}, pages
  269--278, 2020.

\bibitem[Wang et~al.(2018)Wang, Yu, Dou, Darrell, and
  Gonzalez]{wang2018skipnet}
Xin Wang, Fisher Yu, Zi-Yi Dou, Trevor Darrell, and Joseph~E Gonzalez.
\newblock Skipnet: Learning dynamic routing in convolutional networks.
\newblock In \emph{Proceedings of the European Conference on Computer Vision
  (ECCV)}, pages 409--424, 2018.

\bibitem[Wang et~al.(2020)Wang, Zhao, Dai, Fifty, Lin, Hong, and
  Chi]{wang2020small}
Yuyan Wang, Zhe Zhao, Bo~Dai, Christopher Fifty, Dong Lin, Lichan Hong, and
  Ed~H Chi.
\newblock Small towers make big differences.
\newblock \emph{arXiv preprint arXiv:2008.05808}, 2020.

\bibitem[Xiao et~al.(2017)Xiao, Rasul, and Vollgraf]{xiao2017fashion}
Han Xiao, Kashif Rasul, and Roland Vollgraf.
\newblock Fashion-mnist: a novel image dataset for benchmarking machine
  learning algorithms.
\newblock \emph{arXiv preprint arXiv:1708.07747}, 2017.

\bibitem[Xie and Ermon(2019)]{xie2019reparameterizable}
Sang~Michael Xie and Stefano Ermon.
\newblock Reparameterizable subset sampling via continuous relaxations.
\newblock In \emph{International Joint Conference on Artificial Intelligence},
  2019.

\bibitem[Xie et~al.(2020)Xie, Dai, Chen, Dai, Zhao, Zha, Wei, and
  Pfister]{xie2020differentiable}
Yujia Xie, Hanjun Dai, Minshuo Chen, Bo~Dai, Tuo Zhao, Hongyuan Zha, Wei Wei,
  and Tomas Pfister.
\newblock Differentiable top-k with optimal transport.
\newblock \emph{Advances in Neural Information Processing Systems}, 33, 2020.

\bibitem[Zhao et~al.(2019)Zhao, Hong, Wei, Chen, Nath, Andrews, Kumthekar,
  Sathiamoorthy, Yi, and Chi]{zhao2019recommending}
Zhe Zhao, Lichan Hong, Li~Wei, Jilin Chen, Aniruddh Nath, Shawn Andrews, Aditee
  Kumthekar, Maheswaran Sathiamoorthy, Xinyang Yi, and Ed~Chi.
\newblock Recommending what video to watch next: a multitask ranking system.
\newblock In \emph{Proceedings of the 13th ACM Conference on Recommender
  Systems}, pages 43--51, 2019.

\end{thebibliography}

\clearpage

\appendix
\renewcommand\thefigure{\thesection.\arabic{figure}}  
\renewcommand\thetable{\thesection.\arabic{table}}  
\renewcommand\theequation{\thesection.\arabic{equation}}  
\renewcommand{\thesection}{\Alph{section}}

\section{Additional Related Work} \label{sec:appendix_related_work}

\textbf{MTL: } 
In MTL, deep learning-based architectures that perform soft-parameter sharing, i.e., share model parameters partially, are proving to be effective at exploiting both the commonalities and differences among tasks \cite{ruder2017overview}. One flexible architecture for soft-parameter sharing is the multi-gate MoE \citep{ma2018modeling}. We use the multi-gate MoE in our experiments and compare both sparse and dense gates---\cite{ma2018modeling} considered only dense gates. In addition, several works have recently considered gate-like structures for flexible parameter sharing in MTL. For instance, \cite{sun2019adashare,maziarz2019gumbel} give each task the flexibility to use or ignore components inside the neural network. The decisions are modeled using binary random variables, and the corresponding probability distributions are learned using SGD and the Gumbel-softmax trick \citep{jang2016categorical}. This approach is similar to static gating, but it does not support per-example gating. Moreover, the number of nonzeros cannot be directly controlled (in contrast to our gate). 
Our work is also related to \cite{rosenbaum2018routing} who introduced ``routers'' (similar to gates) that can choose which layers or components of layers to activate per-task. The routers in the latter work are not differentiable and require reinforcement learning.

\section{Methodology Details}
\subsection{Proofs} \label{appendix:proofs}
\subsubsection{Proof Proposition 1}
Let $f = \{ f_i \}_{i \in [n]}$. To prove equivalence, we need to establish the following two directions: (I) an optimal solution $(f^{*}, \alpha^{*}, Z^{*})$ to Problem (5) can be used to construct a feasible solution $(f, w)$ to Problem (2) and both solutions have the same objective, and (II) an optimal solution $(f^{*}, w^{*})$ to Problem (2) can be used to construct a feasible solution $(f, \alpha, Z)$ to Problem  (5) and both solutions have the same objective. Direction (I) is trivial: the solution defined by $f = f^{*}$ and $w = q(\alpha^{*}, Z^{*})$ is feasible for Problem (2) and has the same objective as $(f^{*}, \alpha^{*}, Z^{*})$.

Next, we show Direction (II). Let $s^{*} = \| w^{*} \|_0$ and denote by $t_j$ the index of the $j$-th largest element in $w^{*}$, i.e., the nonzero entries in $w^{*}$ are $w_{t_1}^{*} > w_{t_2}^{*} > \dots > w_{t_{s^{*}}}^{*}$. For every $i \in [s^{*}]$, set $z^{(i)}$ to the binary representation of $t_{i} - 1$. If $s^{*} < k$, then we set the remaining (unset) $z^{(i)}$'s as follows: for $i \in \{ s^{*}+1, s^{*}+2, \dots, k \}$ set $z^{(i)}$ to the binary representation of $t_{s^{*}} - 1$. By this construction, the nonzero indices selected by ${r}(z^{(i)})$, $i \in [k]$ are exactly the nonzero indices of $w^{*}$. 

To construct $\alpha$, there are two cases to consider: (i) $s^{*} = k$ and (ii) $s^{*} < k$. If $s^{*} = k$, then set $\alpha_i = \log(w^{*}_{t_i})$ for $i \in [k]$. Therefore, $\sigma(\alpha)_i = w^{*}_{t_i}$ for $i \in [k]$, and consequently $q(\alpha, Z) = w^{*}$. Otherwise, if $s^{*} < k$, then set $\alpha_i = \log(w^{*}_{t_i})$ for $i \in [s^{*} - 1]$ and $\alpha_i =  \log(w^{*}_{t_{s^{*}}}/(k - s^{*}))$ for $i \in [s^{*}, s^{*} + 1, \dots, k]$. Thus, for $i \in [s^{*} - 1]$, we have $\sigma(\alpha)_i = w^{*}_{t_i}$, i.e., the weights of the nonzero indices $t_j$, $j \in [s^{*}-1]$ in $q(\alpha, Z)$ are equal to those in $w^{*}$. The weight assigned to the nonzero index $t_{s^{*}}$ in $q(\alpha, Z)$ is: $\sum_{i \in [s^{*}, s^{*} + 1, \dots, k]} \sigma(\alpha)_i = \sum_{i \in [s^{*}, s^{*} + 1, \dots, k]}  w^{*}_{t_{s^{*}}}/(k - s^{*})  = w^{*}_{t_{s^{*}}}$. Therefore, $q(\alpha, Z) = w^{*}$. In both (i) and (ii), we have $q(\alpha, Z) = w^{*}$, so the solution $(f^{*}, \alpha, Z )$ is feasible and has the same objective as $(f^{*}, w^{*})$.

\subsubsection{Proof of Proposition 2}
First, we will use induction to show that $r(S(z))$ belongs to the probability simplex. Specifically, we will prove that for any integer $t \geq 1$ and $z \in \mathbb{R}^{t}$, $r(S(z))$ belongs to the probability simplex.

Our base case is for $t = 1$. In this case, there is a single binary encoding variable $z_1 \in \mathbb{R}$ and $2$ experts. The single expert selector $r(S(z_1))$ is defined as follows: $r(S(z_1))_1 = 1 - {S}(z_1)$ and $r(S(z_1))_2 = S(z_1)$. The latter two terms are non-negative and sum up to 1. Thus, $r(S(z_1))$ belongs to the probability simplex.

Our induction hypothesis is that for some $t \geq 1$ and any $z \in \mathbb{R}^{t}$, $r(S(z))$ belongs to the probability simplex. For the inductive step, we need to show that for any $v \in \mathbb{R}^{t+1}$, $r(S(v))$ belongs to the probability simplex. From the definition of $r(.)$, the following holds:
\begin{align} \label{eq:rt1}
r(S(v))_i =
\begin{cases}
 r(S([v_1, v_2, \dots, v_t]^T))_i (1 - {S}(v_{t+1})) & i \in [2^{t}] \\
 r(S([v_1, v_2, \dots, v_t]^T))_{i - 2^t} {S}(v_{t+1}) & i \in [2^{t+1}]\setminus[2^t]
\end{cases}
\end{align}
By the induction hypothesis, we have $r(S([v_1, v_2, \dots, v_t]^T))_i \geq 0$ for any $i$. Moreover, $S(.)$ is a non-negative function. Therefore,  $r(S(v))_i \geq 0$ for any $i$. It remains to show that the sum of the entries in $r(S(v))$ is $1$, which we establish next:
\begin{align}
\sum_{i=1}^{2^{t+1}} r(S(v))_i & = \sum_{i=1}^{2^{t}} r(S(v))_i + \sum_{i=2^{t}+1}^{2^{t+1}} r(S(v))_i \nonumber \\ 
& \stackrel{\eqref{eq:rt1}}{=}  \sum_{i=1}^{2^{t}} r(S([v_1, \dots, v_t]^T))_i (1 - {S}(v_{t+1})) + \sum_{i=2^{t}+1}^{2^{t+1}} r(S([v_1, \dots, v_t]^T))_{i - 2^t} {S}(v_{t+1}) \label{eq:proof_change_of_var}
\end{align}
Using a change of variable, the second summation in the above can be rewritten as follows: $\sum_{i=2^{t}+1}^{2^{t+1}} r(S([v_1, v_2, \dots, v_t]^T))_{i - 2^t} {S}(v_{t+1}) = \sum_{i=1}^{2^{t}} r(S([v_1, v_2, \dots, v_t]^T))_{i} {S}(v_{t+1})$. Plugging the latter equality into \eqref{eq:proof_change_of_var} and simplifying, we get $\sum_{i=1}^{2^{t+1}} r(S(v))_i = 1$. Therefore, $r(S(v))$ belongs to the probability simplex, which establishes the inductive step. Finally, we note that $\tilde{q}(\alpha, Z)$ belongs to the probability simplex since it is a convex combination of probability vectors.

\subsection{Extending DSelect-k to arbitrary $n$} \label{sec:arbitrary_n}
Suppose that the number of experts $n$ is not a power of $2$. For the DSelect-k gate $\tilde{q}(\alpha, Z)$ to work in this setting, we need each single expert selector $r$ used by the gate to be able to handle $n$ experts. Next, we discuss how $r$ can handle $n$ when it is not a power of $2$. Let $m$ be the smallest integer so that $n < 2^m$. Then, we treat the problem as if there are $2^m$ experts and use a binary encoding vector $z \in \mathbb{R}^{m}$. For $i \in [n]$, we let entry $r(z)_{i}$ be the weight of expert $i$ in the MoE. Note that the entries $r(z)_{i}$, $i \in \{n+1, n+2, \dots, 2^m\}$, are not associated with any experts. To avoid a situation where $r(z)$ assigns nonzero probability to the latter entries, we add the following penalty to the objective function: 
$
\frac{\xi}{\sum_{i \in [n]} r(z)_i}
$
where $\xi$ is a non-negative parameter used to control the strength of the penalty. This penalty encourages to $r(z)_{i}, i \in [n]$ (i.e., the entries associated with the $n$ experts) to get more probability. In our experiments, we observe that $\sum_{i \in [n]} r(z)_i$ converges to $1$, when $\xi$ is sufficiently large. The penalty described above is part of our TensorFlow implementation of DSelect-k. We also note that there are other potential alternatives to deal with the entries $r(z)_{i}$, $i \in \{n+1, n+2, \dots, 2^m\}$, without adding a penalty to the objective function. For example, one alternative is to randomly assign each of $r(z)_{i}$, $i \in \{n+1, n+2, \dots, 2^m\}$ to one of the $n$ experts.

\subsection{Smooth-step Function} \label{sec:smooth_step}
\begin{figure}[h]
    \centering
    \includegraphics[scale=0.4]{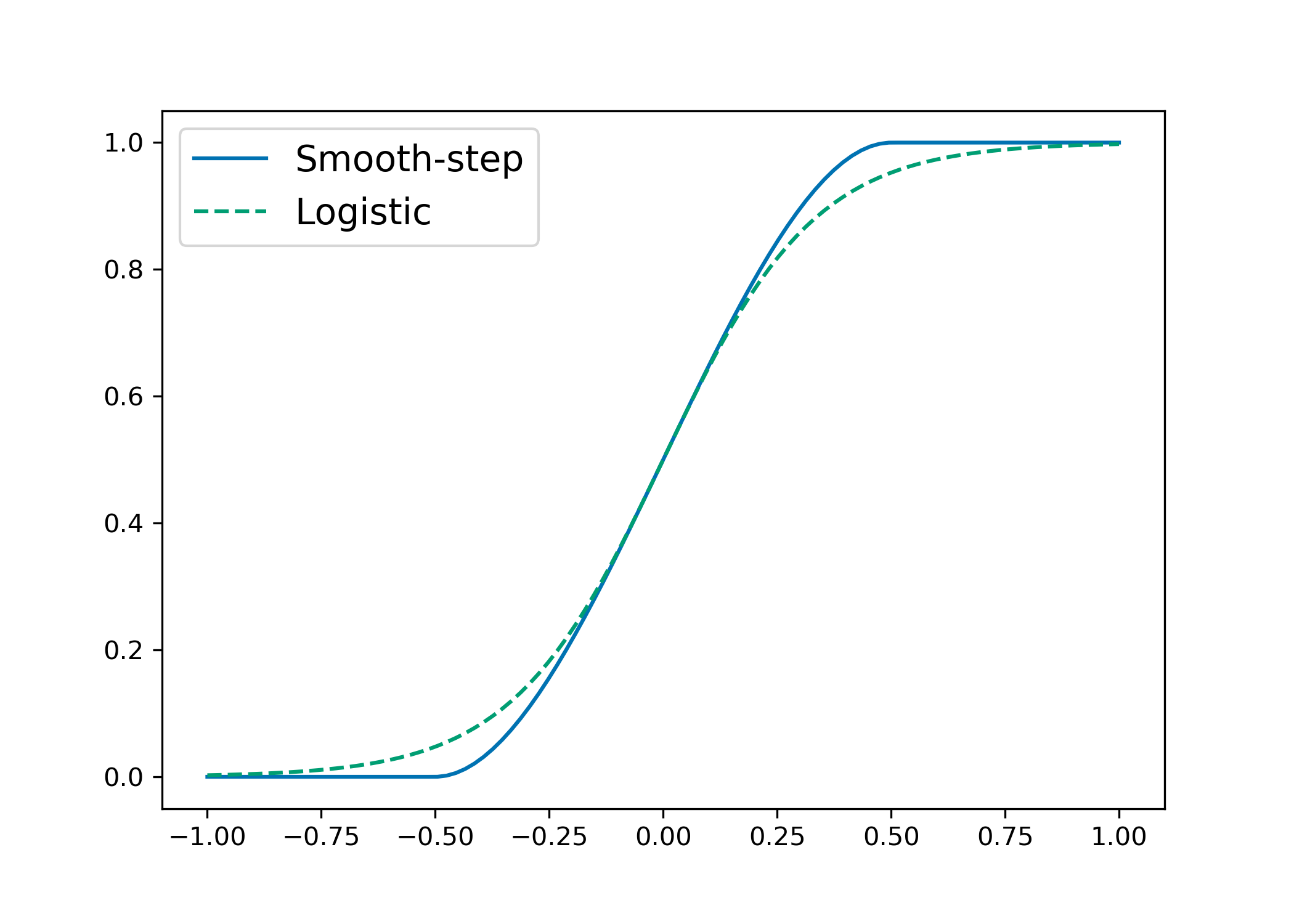}
    \caption{The Smooth-step ($\gamma = 1$) and Logistic functions.}
    \label{fig:smoothstep}
\end{figure}

In Figure \ref{fig:smoothstep}, we plot the smooth-step function \cite{hazimeh2020tree} and the logistic function $L(x) = (1 + e^{-6x})^{-1}$. Note that the logistic function is re-scaled to be on the same scale as the smooth-step function. 

\section{Additional Experimental Results} \label{sec:appendix_experiments}

\subsection{Prediction and Expert Selection Performance on Synthetic Data} \label{sec:synthetic}
In this experiment, we aim to (i) understand how the number of experts and tasks affects the prediction and expert selection performance for the different gates, and (ii) quantify the benefit from binary encoding in our gate through an ablation study. We focus on a static gating setting, where we consider the DSelect-k and Top-k gates, in addition to two variants of the DSelect-k gate used for ablation. To better quantify the expert selection performance and avoid model mis-specification, we use synthetic data generated from a multi-gate  MoE. First, we describe our data generation process.

\textbf{Synthetic Data Generation: } We consider $128$ regression tasks, separated into four mutually exclusive groups: $\{ G_i \}_{i \in [4]}$, where $G_i$ is the set of indices of the tasks in group $i$. As we will discuss next, the tasks are constructed in a way so that tasks within each group are highly related, while tasks across groups are only marginally related. Such a construction mimics real-world applications in which tasks can be clustered in terms of relatedness. 


Each group consists of $16$ tasks which are generated from a group-specific MoE. The group-specific MoE consists of $4$ experts: $\{ f_i \}_{i \in [4]}$. Each expert is the sum of $4$ ReLU-activated units. The output of each task in the group is a convex combination of the $4$ experts. Specifically, for each task $t \in [16]$ in the group, let $\alpha^{(t)} \in \mathbb{R}^{4}$ be a task-specific weight vector. Then, given an input vector $x$, the output of task $t$ is defined as follows:
$$
y^{(t)}(x) := \sum_{i=1}^{4} \sigma(\alpha^{(t)})_i f_i(x)
$$
For each group, we create an instance of the group-specific MoE described above, where we initialize all the weights randomly and independently from the other groups. In particular, we sample the weights of each expert independently from a standard normal distribution. To encourage relatedness among tasks in each group, we sample the task weights $[\alpha^{(1)},\alpha^{(2)} \dots \alpha^{(16)}]$ from a zero-mean multivariate normal distribution where we set the correlation between any two task weights to $0.8$. 

To generate the data, we sample a data matrix $X$, with $140{,}000$ observations and $10$ features, from a standard normal distribution. The data matrix is shared by all $128$ tasks and the regression outputs are obtained by using $X$ as an input to each group-specific MoE. We use  $100{,}000$ observations for the training set and $20{,}000$ observations for each of the validation and testing sets. 

\textbf{Experiment Design: } We consider a multi-gate MoE and compare the following static gates: DSelect-k gate, Top-k gate, and an ``ablation'' gate (which will be discussed later in this section). Our goal is to study how, for each gate, the number of tasks affects the prediction and expert selection performance. To this end, we consider $4$ regression problems, each for a different subset of the $128$ tasks; specifically, we consider predicting the tasks in (i) $G_1$ ($16$ tasks), (ii) $G_1 \cup G_2$ ($32$ tasks),  (iii) $G_1 \cup G_2 \cup G_3$ ($64$ tasks), and (iv) $G_1 \cup G_2 \cup G_3 \cup G_4$ ($128$ tasks). In each of the four problems, we use a multi-gate MoE to predict the outputs of the corresponding tasks simultaneously. The MoE has the same number of experts used to generate the data, i.e., if $T$ is the total number of tasks in the problem, the MoE consists of $T/4$ experts, where the experts are similar to those used in data generation (but are trainable). Each task is associated with a task-specific gate, which chooses a convex combination of $4$ out of the $T/4$ experts. Note that unlike the architecture used to generate the data, each task gate here is connected to all experts, even those belonging to the unrelated groups. The architecture used to generate the data can be recovered if the task gates across groups do not share experts, and the task gates within each group share the same $4$ experts. We use squared error loss for training and tuning.

\begin{figure*}[htb] 
    \centering
    \subfloat{{\includegraphics[width=5.5cm]{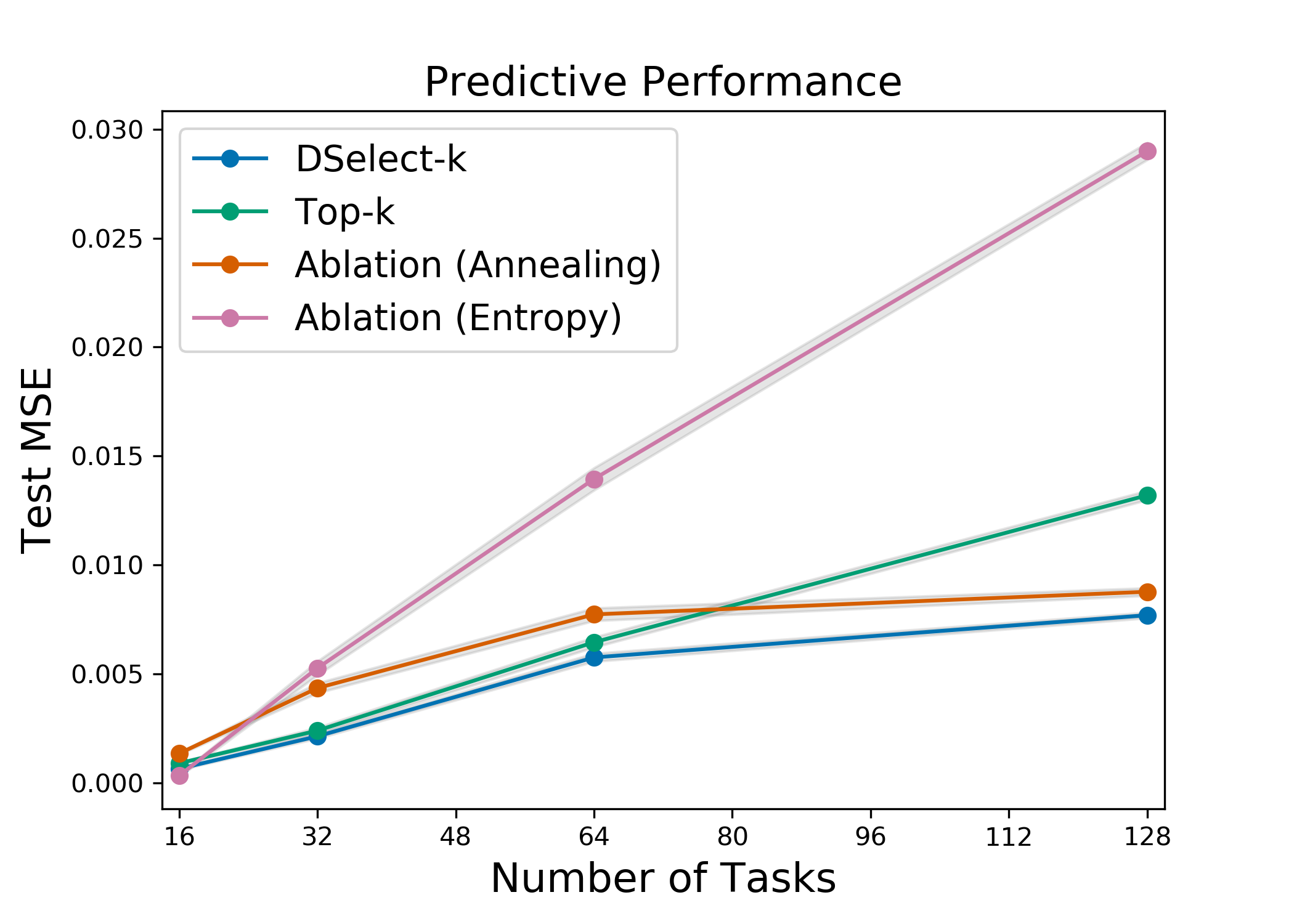}}}%
    \hspace{-0.5cm}
    \subfloat{{\includegraphics[width=5.5cm]{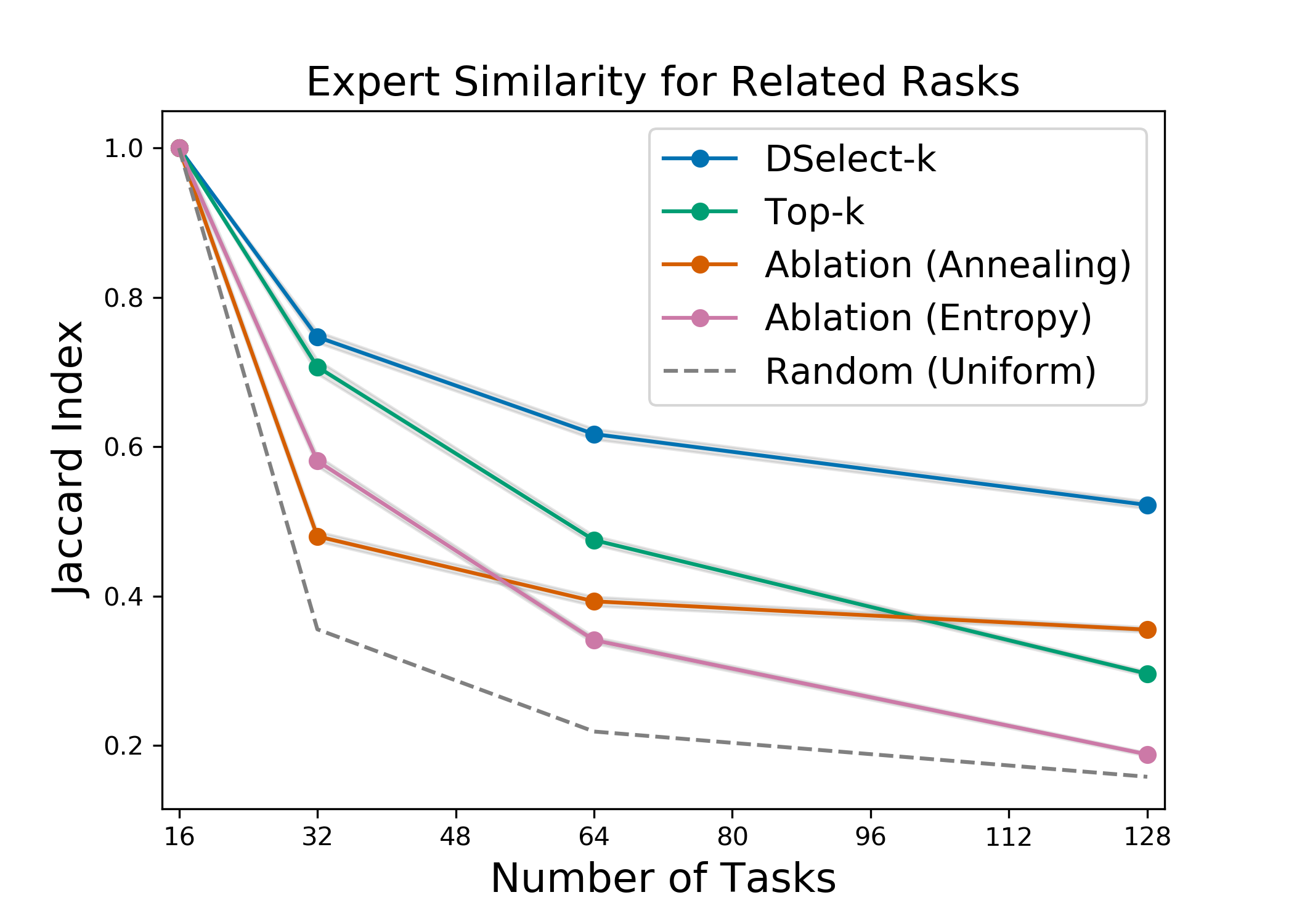} }}%
    \hspace{-0.5cm}
    \subfloat{{\includegraphics[width=5.5cm]{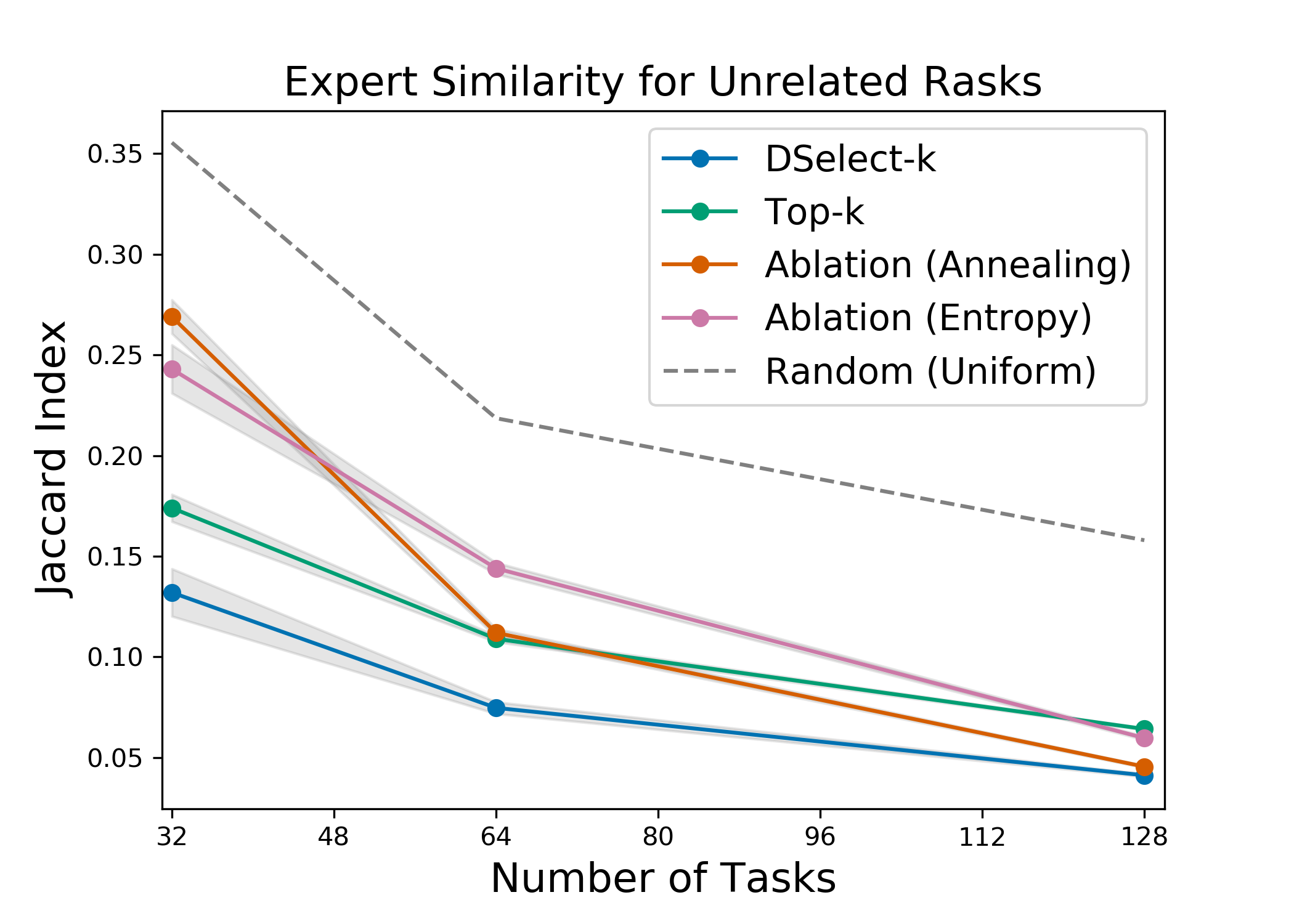} }}%
    \caption{Predictive and expert selection performance on synthetic data generated from a MoE.}%
    \label{fig:synthetic}
\end{figure*}

\textbf{Ablation: } In addition to comparing the DSelect-k and Top-k gates, we perform an ablation study to gain insight on the role of binary encoding in the DSelect-k gate. Recall that in the DSelect-k gate, we introduced the single expert selector which learns a one-hot encoded vector (using a binary encoding scheme). In the literature, a popular way to learn such one-hot encoded vectors is by using a softmax (with additional heuristics such as temperature annealing to ensure that the probability concentrates on one entry). Thus, in our ablation study, we consider the DSelect-k gate $\tilde{q}(\alpha, Z)$, and we replace each single expert selector $r(.)$ with a softmax-based selector. More precisely, let $\alpha \in \mathbb{R}^{k}$ and $\beta^{(i)} \in \mathbb{R}^{n}$, $i \in [k]$, be learnable parameter vectors. Then, we consider the following ``ablation'' gate:
$
  h(\alpha, \beta) :=  \sum_{i=1}^{k} \sigma(\alpha)_i \sigma(\beta^{(i)}).  
$
Here, $\sigma(\alpha)_i$ determines the weight assigned to selector $i$, and $\sigma(\beta^{(i)})$ acts as a surrogate to the  single expert selector $r(S(z^{(i)}))$. To ensure that $\sigma(\beta^{(i)})$ selects a single expert (i.e., leads to a one-hot encoding), we consider two alternatives: (i) annealing the temperature of $\sigma(\beta^{(i)})$ during training\footnote{There are pathological cases where annealing the temperature in softmax will converge to more than one nonzero entry. This can happen when multiple entries in the input to the softmax have exactly the same value.}, and (ii) augmenting the objective with an entropy regularization term (similar to that of the DSelect-k gate) to minimize the entropy of each $\sigma(\beta^{(i)})$. In our results, we refer to (i) by ``Ablation (Annealing)'' and to (ii) by ``Ablation (Entropy)''. Note that these two ablation alternatives can converge to a one-hot encoding asymptotically (due to the nature of the softmax), whereas our proposed gate can converge in a finite number of steps.

\textbf{Measuring Expert Selection Performance: } To quantify the similarity between the experts selected by the different tasks, we use the Jaccard index. Given two tasks, let $A$ and $B$ be the sets of experts selected by the first and second tasks, respectively. The  Jaccard index of these two sets is defined by: $|A \cap B|/|A \cup B|$. In our experiments, we compute: (i) the average Jaccard index for the related tasks, and (ii) the average Jaccard index for unrelated tasks. Specifically, we obtain (i) by computing the Jaccard index for each pair of related tasks, and then averaging. We obtain (ii) by computing the Jaccard index over all pairs of tasks that belong to different groups (i.e., pairs in the same group are ignored), and then averaging.

\textbf{Results:} After tuning, we train each competing model, with the best hyperparameters, for $100$ randomly-initialized repetitions. In Figure \ref{fig:synthetic}, we plot the performance measures (averaged over the repetitions) versus the number of tasks. In the left plot, we report the MSE on the test set. In the middle and right plots we report the (averaged) Jaccard index for the related and unrelated tasks, respectively. In the latter two plots, we also consider a random gate which chooses $4$ experts uniformly at random, and plot the expected value of its Jaccard index. In Figure \ref{fig:synthetic} (middle), a larger index is better since the related tasks will be sharing more experts. In contrast, in Figure \ref{fig:synthetic} (right), a lower index is preferred since the unrelated tasks will be sharing less experts. For all methods, the Jaccard index in Figures \ref{fig:synthetic} (middle) and (right) decreases with the number of tasks. This is intuitive, since as the number of tasks increases, we use more experts, giving any two given gates more flexibility in choosing mutually exclusive subsets of experts.

Overall, the results indicate that DSelect-k gate significantly outperforms Top-k in all the considered performance measures, and the differences become more pronounced as the number of tasks increases. For example, at $128$ tasks, DSelect-k achieves over $40\%$ improvement in  MSE and $76\%$ improvement in Jaccard index for related tasks, compared to Top-k. The DSelect-k gate also outperforms the two ablation gates in which we replace the binary encoding by a Softmax-based selector. The latter improvement suggests that the proposed binary encoding scheme is relatively effective at selecting the right experts. We also investigated the poor performance of the Ablation (Entropy) gate, and it turns out that the Softmax-based single expert selectors, i.e., the $\sigma(\beta^{(i)})$'s, tend to select the same expert. Specifically, we set $k=4$ in the ablation gate, but it ends up selecting $\sim 2$ experts in many of the training repetitions. In contrast, the DSelect-k and Top-k gates select $4$ experts.

\subsection{Gate Visualizations} \label{sec:visualization_appendix}
\subsubsection{MovieLens} 
\begin{figure}[htbp] 
    \centering
    \includegraphics[scale=0.22]{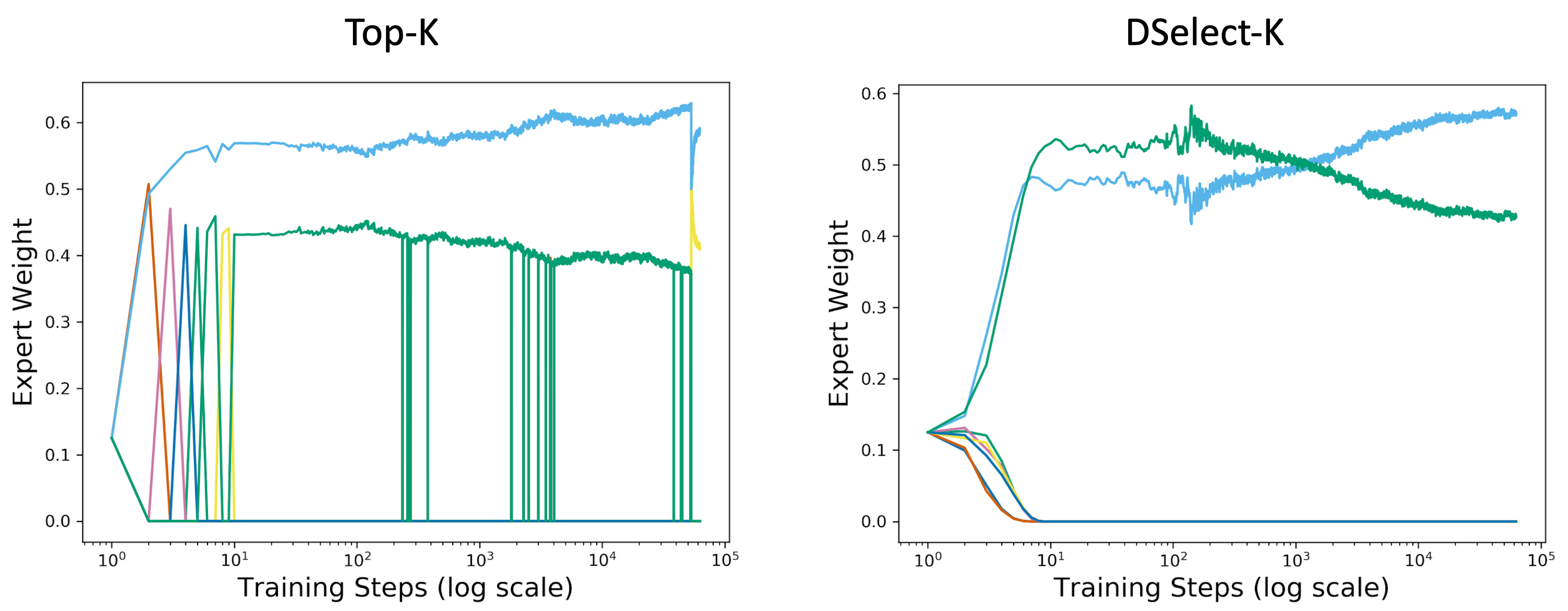}
    \caption{Expert weights during training on the MovieLens dataset. Each color corresponds to a separate expert. The plots are for the best models obtained after tuning.}
    \label{fig:movielens_viz}
\end{figure}

In Figure \ref{fig:movielens_viz}, we plot the expert weights during training on the MovieLens dataset, for the Top-k and DSelect-k gates (after tuning both models). The plots show that Top-k exhibits frequent ``jumps'', where in a single training step an expert's weight can abruptly change from a nonzero value to zero. These jumps keep occurring till the end of training (at around $10^5$ training steps). In contrast, DSelect-k has smooth transitions during training. Additional details on the MovieLens dataset and the MoE architecture used can be found in Section 4.1 of the paper. 

\subsubsection{Synthetic Data} 
\begin{figure}[htbp]
    \centering
    \includegraphics[scale=0.2]{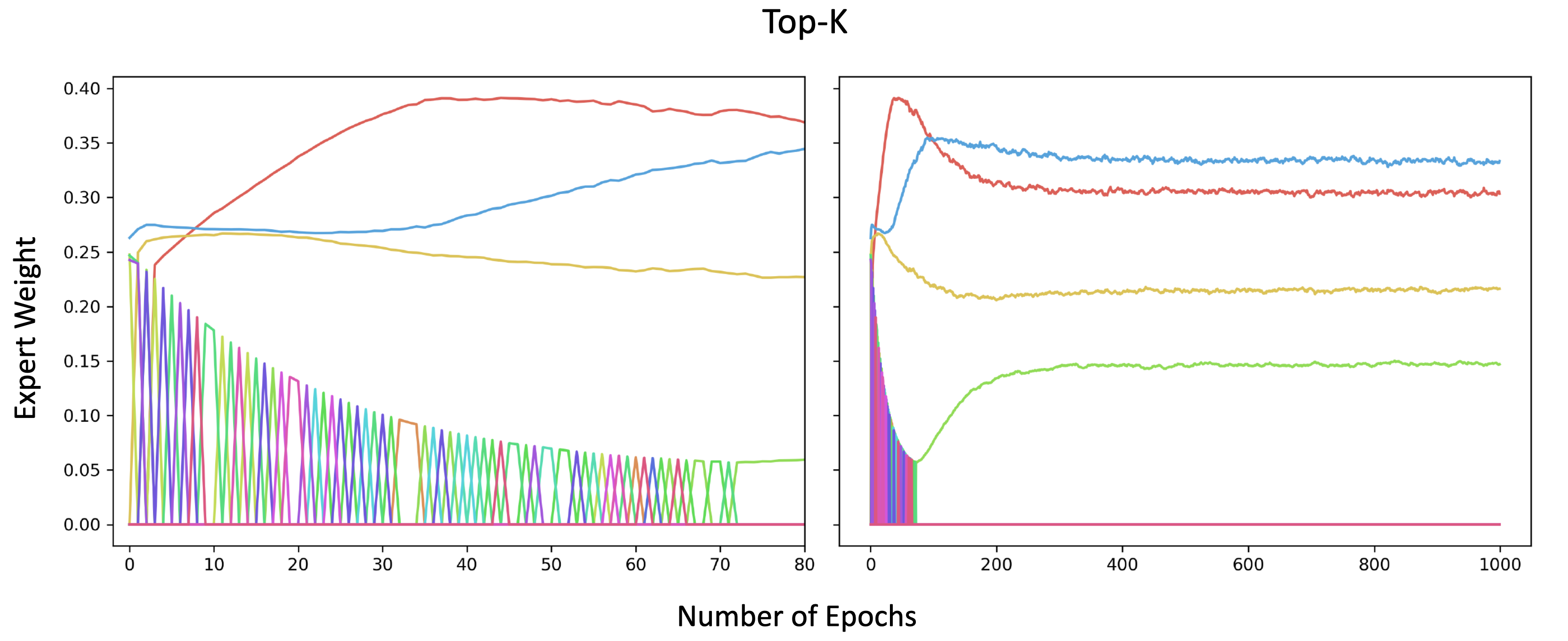}
    \caption{Expert weights during training on synthetic data generated from a MoE. Each color corresponds to a separate expert. The left plot is a magnified version of the right plot. The plots are for the best model obtained after tuning.}
    \label{fig:topk_viz}
\end{figure}

\begin{figure}[htbp]
    \centering
    \includegraphics[scale=0.2]{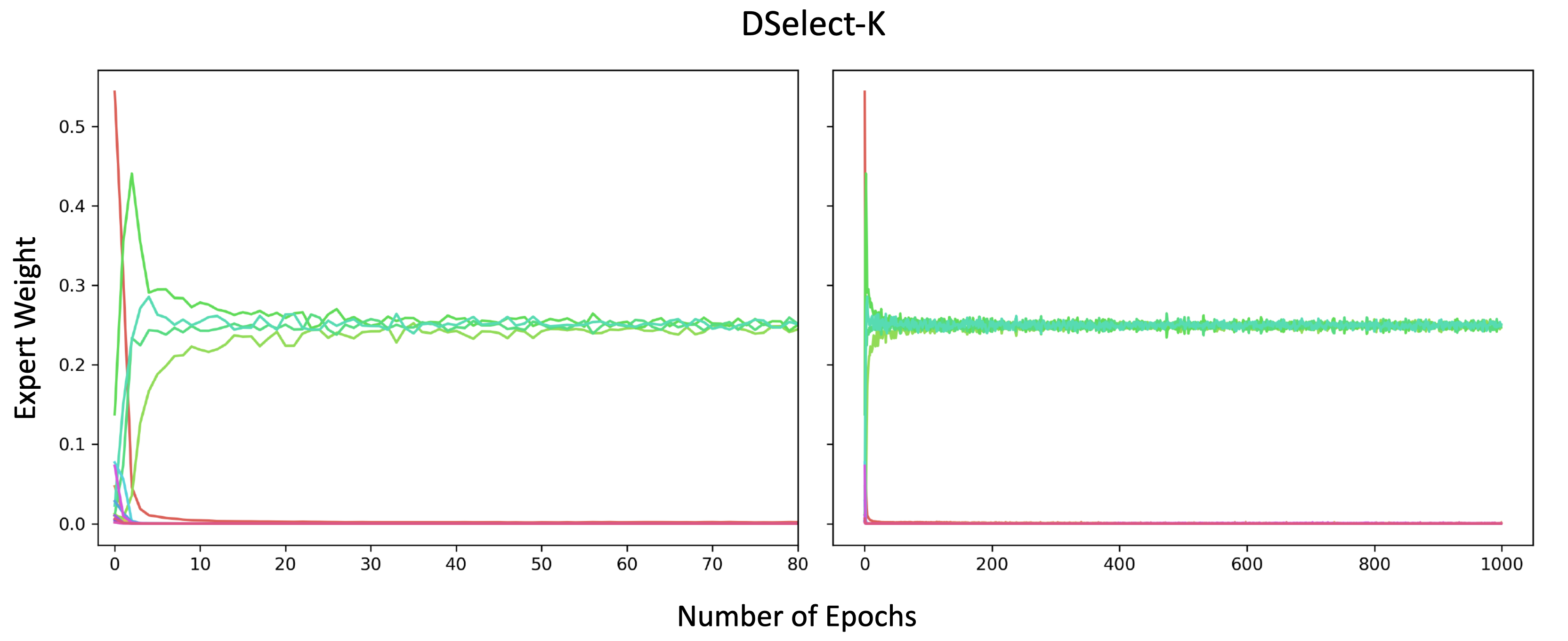}
    \caption{Expert weights during training on synthetic data generated from a MoE. Each color corresponds to a separate expert. The left plot is a magnified version of the right plot. The plots are for the best model obtained after tuning.}
    \label{fig:dselect-k_viz}
\end{figure}

Here we consider a binary classification  dataset generated from a static MoE that consists of $4$ experts. We train another MoE model which employs $16$ experts: $4$ of these experts are copies (i.e., have exactly same weights) of the $4$ experts used in data generation, and the rest of the experts are randomly initialized. We freeze all the experts and train only over the gate parameters. In this simple setting, we expect the gate to be able to recover the $4$ experts that were used to generate the data. We trained two MoE models: one based on Top-k and another based on DSelect-k. After tuning both models, Top-k recovered only $1$ right expert (and made $3$ mistakes), whereas our model recovered all $4$ experts. In Figures \ref{fig:topk_viz} and \ref{fig:dselect-k_viz}, we plot the expert weights during training, for the Top-k and DSelect-k gates, respectively. The Top-k exhibits a sharp oscillatory behavior during training, whereas DSelect-k has smooth transitions.

\textbf{Additional details on data generation and model: }
We consider a randomly initialized ``data-generating'' MoE with $4$ experts (each is a ReLU-activated dense layer with $4$ units). The output of the MoE is obtained by taking the average of the $4$ experts and feeding that into a single logistic unit. We generate a multivariate normal data matrix $X$ with 20,000 observations and 10 features (10,000 observations are allocated for each of the training and validation sets) . To generate binary classification labels, we use $X$ as an input to the data-generating MoE and apply a sign function to the corresponding output. For training and tuning, we consider a MoE architecture with $16$ experts: $4$ of these experts are copies of the experts used in data generation, and the rest of the experts are initialized randomly. All experts are frozen (not trainable). A trainable gate chooses $4$ out of the $16$ experts, and the final result is fed into a single logistic unit. We optimized the cross-entropy loss using Adam with a batch size of $256$, and tuned over the learning rate in $\{10^{-1}, 10^{-2}, \dots, 10^{-5}\}$. 

\subsection{Gate Convergence and FLOPS} \label{sec:convergence_appendix}
In Table \ref{table:convergence}, we report the percentage of training steps required for $S(Z)$ to converge to a binary matrix in the DSelect-k gate, on several real datasets. These results are based on the tuned models discussed in Section 4 of the paper. We also report the number of floating point operations (FLOPS) required by the MoE based on DSelect-k relative to the MoE based on Top-k, during training. The results indicate that the number of training steps till convergence to a binary matrix depends on the specific dataset: ranging from only $0.04\%$ on the MovieLens dataset to $80\%$ on the Multi-Fashion MNIST. Moreover, on certain datasets (MovieLens with $\alpha=0.9$ and Multi-MNIST), DSelect-k requires less FLOPS during training than Top-k, i.e., DSelect-k is effective at conditional training (on these particular datasets). 
\begin{table}[htbp]
\caption{\small{We report two statistics: (i) Percentage of training steps required for the DSelect-k gate to converge to a binary matrix, and (ii) the number of FLOPS needed by the DSelect-k based MoE during training relative to that of Top-k. The parameter $\alpha$ controls the weight assigned to task 1's loss in the MovieLens dataset---see Section 4.1 of the paper for more details.}}
\label{table:convergence}
\centering
\resizebox{0.85\columnwidth}{!}{
\begin{tabular}{l|ccc|c|c|}
\cline{2-6}
                             & \multicolumn{3}{c|}{MovieLens}             & \multirow{2}{*}{Multi-MNIST} & \multirow{2}{*}{Multi-Fashion} \\
                             & $\alpha=0.1$ & $\alpha=0.5$ & $\alpha=0.9$ &                              &                                      \\ \hline
\multicolumn{1}{|l|}{\% Training Steps until Binary $S(Z)$} & 11.37 &	9.39 & 0.04 & 42.33 & 80.02  \\
\multicolumn{1}{|l|}{FLOPS(DSelect-k)/FLOPS(Top-k)} & 1.5	& 1.2 &	0.6 & 0.8 &	1.2          \\ \hline
\end{tabular}
}
\end{table}

\subsection{MovieLens} \label{sec:movie_lens_extra_table} In Table \ref{table:movie_lens_extra_table}, we report the accuracy for task 1 (classification) and the loss for task 2 ( regression) for the competing methods on the MovieLens dataset.
\begin{table*}[htbp]
\centering
\caption{\small{Mean test loss for task 2 (T2) and accuracy for task 1 (T1) on MovieLens (standard error is shown next to each mean). The parameter $\alpha$ determines the weight of Task 1's loss (see text for details). The test loss is  multiplied by $10^4$.}}
\label{table:movie_lens_extra_table}
\resizebox{0.85\columnwidth}{!}{
\begin{tabular}{c|l|cc|cc|cc|}
\cline{3-8}
\multicolumn{1}{l}{}                                                                                & \multirow{2}{*}{} & \multicolumn{2}{c|}{$\alpha = 0.1$} & \multicolumn{2}{c|}{$\alpha = 0.5$} & \multicolumn{2}{c|}{$\alpha = 0.9$} \\
\multicolumn{1}{l}{}                                                                                &                         & T2 Loss                & T1 Accuracy       & T2 Loss                 & T1 Accuracy       & T2 Loss                   & T1 Accuracy     \\ \hline
\multicolumn{1}{|c|}{\multirow{3}{*}{\begin{tabular}[c]{@{}c@{}}Static\end{tabular}}}      & DSelect-k           & $4038 \pm 5$	&	$83.03 \pm 0.07$	&	$3926 \pm 6$	&	$83.86 \pm 0.02$	&	$3943 \pm 5$	&	$84.04 \pm 0.01$	\\
\multicolumn{1}{|c|}{}                                                                               & Top-k       &   $4056 \pm 4$	&	$84.09 \pm 0.05$	&	$4002 \pm 4$	&	$84.21 \pm 0.02$	&	$3884 \pm 4$	&	$84.17 \pm 0.02$	\\
\multicolumn{1}{|c|}{}                                                                               & Gumbel Softmax & $4172 \pm 2$	&	$80.76 \pm 0.04$	&	$4085 \pm 3$	&	$83.71 \pm 0.01$	&	$3878 \pm 4$	&	$84.17 \pm 0.02$	\\ \hline
\multicolumn{1}{|c|}{\multirow{2}{*}{\begin{tabular}[c]{@{}c@{}}Per-example\end{tabular}}} & DSelect-k      & $4030 \pm 7$	&	$83.03 \pm 0.12$	&	$3981 \pm 7$	&	$83.92 \pm 0.03$	&	$3932 \pm 1$	&	$84.07 \pm 0.02$	\\
\multicolumn{1}{|c|}{}                                                                               & Top-k     & $4057 \pm 9$	&	$83.28 \pm 0.1$	&	$3995 \pm 8$	&	$83.91 \pm 0.03$	&	$3914 \pm 4$	&	$84.05 \pm 0.01$	\\ \hline
\multicolumn{1}{|c|}{\multirow{2}{*}{Baselines}}                                                     & Softmax MoE                 & $4047 \pm 1$	&	$78.73 \pm 0.01$	&	$4028 \pm 3$	&	$83.56 \pm 0.01$	&	$3970 \pm 3$	&	$83.86 \pm 0.02$	\\
\multicolumn{1}{|c|}{}                                                                               & Shared Bottom           & $3993 \pm 2$	&	$79.06 \pm 0.02$	&	$3875 \pm 2$	&	$82.65 \pm 0.02$	&	$3991 \pm 2$	&	$84.01 \pm 0.01$	\\ \hline
\end{tabular}
}
\end{table*}
\section{Experimental Details} \label{sec:experimental_details}

\textbf{Computing Setup:} We ran the experiments on a cluster that automatically allocates the computing resources. We do not report the exact specifications of the cluster for confidentiality.

\textbf{Gumbel-softmax Gate:} \cite{sun2019adashare,maziarz2019gumbel} present an approach for learning which layers in a neural network to activate on a per-task basis. The decision to select each layer is modeled using a binary random variable whose distribution is learned using the Gumbel-softmax trick. Note that the latter approach does not consider a MoE model. Here we adapt the latter approach to the  MoE; specifically we consider a Gumbel-softmax gate that uses binary variables to determine which experts to select. Given $n$ experts $\{f_i\}_{i=1}^{n}$, this gate uses $n$ binary random variables $\{U_i\}_{i=1}^{n}$, where $U_i$ determines whether expert $f_i$ is selected. Moreover, the gate uses an additional learnable vector $\alpha \in \mathbb{R}^{n}$ that determines the weights of the experts. Specifically, the gate is a function $d(\alpha, U)$ whose $i$-th component (for any $i \in [n])$ is given by:
$$
d(\alpha, U)_i = \sigma(\alpha)_i  U_i 
$$
To learn the distribution of the $U_i$'s, we use the Gumbel-softmax trick as described in \cite{sun2019adashare}. Moreover, following  \cite{sun2019adashare}, we add the following sparsity-inducing penalty to the objective function: $\lambda \sum_{i \in [n]} \log{\psi_i}$, where $\psi_i$ is the Bernoulli distribution parameter of $U_i$, and $\lambda$ is a non-negative parameter used to control the number of nonzeros selected by the gate.  Note that the latter penalty cannot directly control the number of nonzeros as in DSelect-k or Top-k. 

\subsection{MovieLens} \label{sec:movie_lens_extra_details}

\textbf{Architecture: } For MoE-based models, we consider a multi-gate MoE architecture (see Figure 1), where each task is associated with a separate gate. The MoE uses $8$ experts, each of which is a ReLU-activated dense layer with $256$ units, followed by a dropout layer (with a dropout rate of $0.5$). For each of the two tasks, the corresponding  convex combination of the experts is fed into a task-specific subnetwork. The subnetwork is composed of a dense layer (ReLU-activated with $256$ units) followed by a single unit that generates the final output of the task. The shared bottom model uses a dense layer (whose number of units is a hyperparameter) that is shared by the two tasks, followed by a dropout layer (with a rate of $0.5$). For each task, the output of the shared layer is fed into a task-specific subnetwork (same as that of the MoE-based models).

\textbf{Hyperparameters and Tuning: } We tuned each model using random grid search, with an average of $5$ trials per grid point. We used Adagrad with a batch size of $128$ and considered the following hyperparameters and ranges: Learning Rate: $\{0.001, 0.01, 0.1, 0.2, 0.3 \}$, Epochs: $\{5, 10, 20, 30, 40, 50 \}$, k for Top-k and DSelect-k: $\{2, 4 \}$, $\lambda$ for DSelect-k: $\{0.1, 1, 10 \}$, $\gamma$ for smooth-step: $\{1, 10\}$, Units in Shared bottom: $\{ 32, 256, 2048, 4096, 8192 \}$, $\lambda$ in Gumbel-softmax: $\{10^{-6}, 10^{-5}, \dots,  10 \}$. For Gumbel-softmax, we pick the best solution whose expected number of nonzeros is less than or equal to $4$.

\subsection{Multi-MNIST and Multi-Fashion MNIST} \label{sec:multi_mnist_extra_details}

\textbf{Architecture: } MoE-based models use a multi-gate MoE (as in Figure 1). Each of the $8$ experts is a CNN that is composed (in order) of: (i) convolutional layer 1 (kernel size = 5, number of filters = 10, ReLU-activated) followed by max pooling, (ii) convolutional layer 2 (kernel size = 5, number of filters = 20, ReLU-activated) followed by max pooling, and (iii) a stack of ReLU-activated dense layers with $50$ units each (the number of layers is a hyperparameter). The subnetwork specific to each task is composed of a stack of 3 dense layers: the first two have 50 ReLU-activated units and the third has 10 units followed by a softmax. The shared bottom model uses a shared CNN (with the same architecture as the CNN in the MoE). For each task, the output of the shared CNN is fed into a task-specific subnetwork (same as that of the MoE-based models).

\textbf{Hyperparameters and Tuning: } We tuned each model using random grid search, with an average of $5$ trials per grid point. We used Adam with a batch size of $256$ and considered the following hyperparameters and ranges: Learning Rate: $\{0.01, 0.001, 0.0001, 0.00001 \}$, Epochs: $\{25, 50, 75, 100 \}$, k for Top-k and DSelect-k: $\{2, 4 \}$, $\gamma$ for smooth-step: $\{0.1, 1, 10\}$, Number of dense layers in CNN: $\{ 1, 3, 5 \}$, $\lambda$ in Gumbel-softmax: $\{0, 10^{-3}, 10^{-2}, 10^{-1}, 1, 10, 1000 \}$. For Gumbel-softmax, we pick the best solution whose expected number of nonzeros is less than or equal to $4$.

\subsection{Recommender System} \label{sec:recommender_extra_details}
Each of the $8$ experts in the MoE consists of a stack of ReLU-activated dense layers with $256$ units each. We fix k to 2 in both DSelect-k and Top-k. We tune over the learning rate and architecture. For both models, training is terminated when there is no significant improvement in the validation loss.

\subsection{Synthetic Data}
We tuned each model using random grid search, with an average of $5$ trials per grid point. We used Adam with a batch size of $256$ and considered the following hyperparameters and ranges:
Learning rate: $\{0.001, 0.01, 0.1\}$, Epochs $\{ 25, 50, 75, 100 \}$, $\gamma$ for smooth-step: $\{ 5, 10, 15 \}$, $\lambda$ for DSelect-k: $\{0.001, 0.005, 0.01, 0.1 \}$, $\lambda$ for Ablation (Entropy): $\{ 10^{-6}, 10^{-5}, 10^{-4}, 10^{-3}, 10^{-2}, 10^{-1}, 1, 100 \}$. Moreover, for Ablation (Annealing), we anneal the temperature of softmax starting from a hyperparameter $s$ down to $10^{-16}$ (the temperatures are evenly spaced on a logarithmic scale). We tune the starting temperature $s$ over $\{10^{-6}, 10^{-5}, 5 \times 10^{-5}, 10^{-4}, 2.5 \times 10^{-4}, 5 \times 10^{-4}, 7.5 \times 10^{-4}, 10^{-3}, 5 \times 10^{-3}, 10^{-2} \}$ (note that such a fine grid was necessary to get annealing to work for the ablation gate).





\end{document}